\documentclass[journal]{IEEEtran}
\ifCLASSINFOpdf
  \usepackage[pdftex]{graphicx}
\else
\fi
\usepackage{amssymb}
\usepackage[cmex10]{amsmath}
\newtheorem{thm}{Theorem}

\usepackage{algorithm}
\usepackage{algorithmic}
\usepackage{cases}
\usepackage{array}
\usepackage[tight,footnotesize]{subfigure}
\begin{document}
%\title{Bare Demo of IEEEtran.cls for Journals}
\title{Large Margin Multi-modal Multi-task Feature Extraction for Image Classification}

%\author{Michael~Shell,~\IEEEmembership{Member,~IEEE,}
%        John~Doe,~\IEEEmembership{Fellow,~OSA,}
%        and~Jane~Doe,~\IEEEmembership{Life~Fellow,~IEEE}% <-this % stops a space
%\thanks{M. Shell is with the Department
%of Electrical and Computer Engineering, Georgia Institute of Technology, Atlanta,
%\thanks{Manuscript received April 19, 2005; revised September 17, 2014.}}

\author{Yong~Luo,
        Yonggang~Wen,~\IEEEmembership{Senior Member,~IEEE,}
        Dacheng~Tao,~\IEEEmembership{Fellow,~IEEE,} \\
        Jie~Gui,~\IEEEmembership{Member,~IEEE,}
        and~Chao~Xu,~\IEEEmembership{Member,~IEEE}% <-this % stops a space
\thanks{Y. Luo and C. Xu are with the Key Laboratory of Machine Perception (Ministry of Education), School of Electronics Engineering and Computer Science, Peking University, Beijing, China. Y. Luo is also with the Division of Networks and Distributed Systems, School of Computer Engineering, Nanyang Technological University, Singapore, and the Centre for Quantum Computation \& Intelligent Systems and the Faculty of Engineering \& Information Technology, University of Technology, Sydney, Sydney, Australia (email: yluo180@gmail.com, xuchao@cis.pku.edu.cn).}% <-this % stops a space
\thanks{Y. Wen is with the Division of Networks and Distributed Systems, School of Computer Engineering, Nanyang Technological University, Singapore (email: ygwen@ntu.edu.sg).}
\thanks{D. Tao is with the Centre for Quantum Computation \& Intelligent Systems and the Faculty of Engineering and Information Technology, University of Technology, Sydney, 81 Broadway Street, Ultimo, NSW 2007, Australia (email: dacheng.tao@uts.edu.au).}% <-this % stops a space
\thanks{J. Gui is with the Hefei Institute of Intelligent Machines, Chinese Academy of Sciences, Hefei 230031, China, and also with the Center for Research on Intelligent Perception and Computing, National Laboratory of Pattern Recognition, Institute of Automation, Chinese Academy of Sciences, Beijing 100190, China (e-mail: guijie@ustc.edu).}
\thanks{\copyright 2015 IEEE. Personal use of this material is permitted. Permission from IEEE must be obtained for all other uses, in any current or future media, including reprinting/republishing this material for advertising or promotional purposes, creating new collective works, for resale or redistribution to servers or lists, or reuse of any copyrighted component of this work in other works.}
}


%\markboth{Journal of \LaTeX\ Class Files,~Vol.~13, No.~9, September~2014}%
%{Shell \MakeLowercase{\textit{et al.}}: Bare Demo of IEEEtran.cls for Journals}

\markboth{$>$ \normalsize{TIP-13616-2015 R}\footnotesize{evision} \normalsize{1} $<$}%
{Shell \MakeLowercase{\textit{et al.}}: Bare Demo of IEEEtran.cls for Journals}




\maketitle

\begin{abstract}
%The abstract goes here.
The features used in many image analysis-based applications are frequently of very high dimension. Feature extraction offers several advantages in high-dimensional cases, and many recent studies have used multi-task feature extraction approaches, which often outperform single-task feature extraction approaches. However, most of these methods are limited in that they only consider data represented by a single type of feature, even though features usually represent images from multiple modalities. We therefore propose a novel large margin multi-modal multi-task feature extraction (LM3FE) framework for handling multi-modal features for image classification. In particular, LM3FE simultaneously learns the feature extraction matrix for each modality and the modality combination coefficients. In this way, LM3FE not only handles correlated and noisy features, but also utilizes the complementarity of different modalities to further help reduce feature redundancy in each modality. The large margin principle employed also helps to extract strongly predictive features so that they are more suitable for prediction (e.g., classification). An alternating algorithm is developed for problem optimization and each sub-problem can be efficiently solved. Experiments on two challenging real-world image datasets demonstrate the effectiveness and superiority of the proposed method.
\end{abstract}


% \fi
\IEEEpeerreviewmaketitle

\section{Introduction}
\label{sec:Introduction}
%\IEEEPARstart{T}{his} demo file is intended to serve as a ``starter file''
%for IEEE journal papers produced under \LaTeX\ using
%IEEEtran.cls version 1.8a and later.
%% You must have at least 2 lines in the paragraph with the drop letter
%% (should never be an issue)
%I wish you the best of success.
%\hfill September 17, 2014

\IEEEPARstart{I}{mage} classification \cite{U-Srinivas-et-al-TIP-2015} lies at the heart of many image analysis-based applications, such as face recognition, web image browsing, and medical and remote sensing image analysis. In these applications, very high-dimensional feature vectors are usually used to represent the images. For example, some widely-used global (such as GIST \cite{A-Oliva-and-A-Torralba-IJCV-2001}) and local (such as SIFT \cite{DG-Lowe-IJCV-2004}) descriptors can have dimensionality of up to several hundred or thousand when used for image classification \cite{M-Guillaumin-et-al-CVPR-2010}.

Feature extraction \cite{I-Guyon-et-al-Book-2006} is therefore a useful tool for removing irrelevant or redundant information and reducing feature dimensionality. It makes the learning process more efficient, reduces the chance of over-fitting, and improves the generalizability of the model \cite{FP-Nie-et-al-NIPS-2010, ZC-Li-et-al-AAAI-2012}. Feature selection and feature transformation are the two main approaches used for feature extraction; in the former, a subset of features is selected from the original, while in the latter the original feature is transformed into a new feature space. The former is often the preferred method \cite{M-Masaeli-et-al-ICML-2010}. Although traditional feature selection methods usually select features from a single task \cite{PS-Bradley-and-OL-Mangasarian-ICML-1998, RO-Duda-et-al-Book-2001, XF-He-et-al-NIPS-2005}, there has recently been a focus on joint feature selection across multiple related tasks \cite{G-Obozinski-et-al-ICMLw-2006, FP-Nie-et-al-NIPS-2010, Y-Yang-et-al-IJCAI-2011, ZC-Li-et-al-AAAI-2012, JL-Tang-et-al-SDM-2013}. This is because joint feature selection exploits task correlations in order to establish the importance of features, and this approach has been empirically demonstrated to be superior to feature selection on separate tasks \cite{G-Obozinski-et-al-ICMLw-2006}. Since all the joint selection methods aim to learn feature selection matrices, which can also be used for feature transformation, we regard these as feature extraction methods in this paper.

Multi-task feature extraction can effectively handle correlated and noisy features but it is not suitable for image classification problems that contain images represented by multi-modal features. Although all the features can be concatenated into a long vector, this strategy ignores the diversity between features and may lead to a severe ``curse of dimensionality''. One recent work considers unsupervised multi-modal feature extraction \cite{JL-Tang-et-al-SDM-2013} based on spectral analysis \cite{UV-Luxburg-SC-2007}, but this method is unsuitable for multi-task feature extraction in the supervised scenario considered in this paper. A supervised multi-modal multi-task feature learning algorithm is proposed in \cite{H-Wang-et-al-ICML-2013}, but the learned features do not have strong prediction power and thus are not appropriate for classification. The tensor-based multi-modal feature selection method presented in \cite{BK-Cao-et-al-ICDM-2014} utilizes the strongly predictive model SVM (support vector machine) to eliminate features, but disregards the relationships between tasks (classes), and the computational complexity is very high.

To overcome the limitations of existing methods, this paper presents a novel large margin multi-modal multi-task feature extraction (LM3FE) algorithm that effectively explores the complementary nature of different modalities obtained from multiple tasks. In particular, LM3FE learns a projection matrix for each modality that transforms the data from the original feature space to a latent feature space. In the latent space, a weighted combination of all the transformed features is then used to predict the ground-truth labels of the training data. The $l_{2,1}$-norm constraints on the projection matrices make them suitable for both feature selection and feature transformation. The prediction loss is chosen as the hinge loss for classification, and the margin maximization principle enhances the prediction power of the selected or transformed features. All the projection matrices of different modalities, the combination weights, and the prediction matrix are learned as a single optimization problem. This approach exploits both the task relationship and the complementary nature of different modalities for effective and strongly predictive feature extraction.

We thoroughly evaluate the proposed LM3FE algorithm on a real-world web image dataset, NUS-WIDE (NUS) \cite{TS-Chua-et-al-CIVR-2009}, and a challenge social image dataset, MIR Flickr (MIR) \cite{MJ-Huiskes-and-MS-Lew-ICMR-2008}. Extensive experiments demonstrate the superiority of the proposed large margin approach by comparing it with other supervised multi-task feature extraction algorithms with feature concatenation (MTFS \cite{G-Obozinski-et-al-ICMLw-2006} and RFS \cite{FP-Nie-et-al-NIPS-2010}), and a competitive multi-modal multi-task feature learning method \cite{H-Wang-et-al-ICML-2013}, as well as a recently proposed multi-modal feature selection algorithm \cite{BK-Cao-et-al-ICDM-2014}.

\section{Related Work}
\label{sec:Related_Work}

Our work is a multi-modal extension of the multi-task feature extraction methods.

\subsection{Multi-task feature extraction}

Dozens of feature extraction (selection or transformation) methods have been proposed in the literature due to the significance of this technique in pattern recognition and machine learning \cite{I-Guyon-et-al-Book-2006, FP-Nie-et-al-AAAI-2008, X-Cai-et-al-ICDM-2011, SM-Xiang-et-al-TNNLS-2012, CP-Hou-et-al-TOC-2014}. We refer to \cite{I-Guyon-et-al-Book-2006} for a survey of traditional feature extraction approaches, and only focus in this paper on joint feature extraction across multiple tasks. An early work of this kind was done by Obozinski et al. \cite{G-Obozinski-et-al-ICMLw-2006}, in which the $l_{2,1}$-norm was introduced to encourage similar sparsity patterns for related tasks in feature selection. This was extended in \cite{FP-Nie-et-al-NIPS-2010} by emphasizing $l_{2,1}$-norm on both the loss and regularization term for the sake of efficiency and robustness. Considering that labeled data might not be available, Yang et al. developed an unsupervised feature selection method \cite{Y-Yang-et-al-IJCAI-2011}, in which feature correlation is exploited via the $l_{2,1}$-norm, and discriminative information is incorporated in learning by the defined local discriminative score. The discriminative information is also exploited in \cite{ZC-Li-et-al-AAAI-2012} for unsupervised feature selection by making use of the spectral clustering to learn pseudo class labels. We differ from them in that multiple types of features are utilized.

\subsection{Multi-modal learning}

%The multi-view learning we refer to here combines the multiple feature representations of an object.
%The multiple views can be the different viewpoints of an object in the camera \cite{T-Maugey-et-al-TIP-2015}, or the various descriptions of a given sample \cite{S-Eleftheriadis-et-al-TIP-2015, J-Woo-et-al-TIP-2015}. We focus on the latter in this paper, and the goal is to fuse the multiple feature representations of an object. We classify the multi-view learning algorithms into three families:
The multiple modalities we refer to here are the various descriptions of a given sample \cite{S-Eleftheriadis-et-al-TIP-2015, J-Woo-et-al-TIP-2015}, and our goal is to fuse the multiple feature representations of an object. We classify the multi-modal learning algorithms into three families:
\begin{itemize}
  \item \textbf{Weighted modality combination:} most multiple kernel algorithms belong to this family. For example, the kernels built on different feature sets were weighted combined in \cite{GRG-Lanckriet-et-al-JMLR-2004} for protein prediction. McFee and Lanckriet \cite{B-McFee-and-G-Lanckriet-JMLR-2011} presented a partial order embedding algorithm to induce a unified similarity space by weighted combining multiple kernels. In addition, a weighted combination of multiple graph Laplacians was proposed in \cite{T-Xia-et-al-TSMCB-2010} for multi-modal spectral embedding.
  \item \textbf{Multi-modal subspace learning:} canonical correlation analysis (CCA) \cite{DR-Hardoon-et-al-NCn-2004} is a highly representative work of this family. The given two modalities are transformed into a subspace where they are maximally correlated. In \cite{M-White-et-al-NIPS-2012}, a convex formulation was proposed for multi-modal subspace learning with conditional independence constraints.
  \item \textbf{Modality agreement exploration:} most of the methods in this family aim to find agreement of different modalities on the unlabeled data, and thus are semi-supervised \cite{A-Blum-and-T-Mitchell-COLT-1998, U-Brefeld-and-T-Scheffer-ICML-2004} or unsupervised \cite{A-Kumar-et-al-NIPS-2011}. For example, the co-training framework presented in \cite{A-Blum-and-T-Mitchell-COLT-1998} utilizes the classifiers trained on two modalities to classify the unlabeled samples and then selects the classified samples with high confidence to boost the classification performance. A co-regularization strategy was presented in \cite{A-Kumar-et-al-NIPS-2011} to maximize the clustering agreement of different modalities for multi-modal spectral clustering.
\end{itemize}

The proposed LM3FE belongs to multi-modal subspace learning but utilizes a weighted modality combination training strategy. Recently, a multi-modal feature selection method \cite{BK-Cao-et-al-ICDM-2014} has been proposed that explores the correlations between different modalities by taking the tensor product of their feature spaces. However, this method cannot handle the multi-class problem naturally and must train an SVM classifier to eliminate one feature at a time. The relationships between different classes are therefore discarded, and the training cost is very high. The closest work to our method is the multi-modal feature learning approach presented in \cite{H-Wang-et-al-ICML-2013} since it also utilizes the $l_{2,1}$-norm to discover the task relationships, but it differs from our method in that the group $l_1$-norm is employed to capture the correlations between modalities. The main drawback of this approach is that the feature weight matrices of different modalities are concatenated and directly utilized as the prediction matrix. Also, least squares loss is adopted, therefore the prediction (e.g., classification) power of the learned features is limited. In the proposed LM3FE, a prediction matrix is learned in addition to the feature extraction matrices, and strongly predictive features are obtained by minimizing the hinge loss under the maximum margin principle.

%\subsection{Subsection Heading Here}
%Subsection text here.

%\IEEEpubidadjcol

%\subsubsection{Subsubsection Heading Here}
%Subsubsection text here.

\section{Multi-task Feature Extraction}
\label{sec:MTFE}

We first briefly introduce the multi-task feature selection (MTFS) framework \cite{G-Obozinski-et-al-ICMLw-2006} and its efficient and robust version \cite{FP-Nie-et-al-NIPS-2010}, both of which can be regarded as feature extraction methods. Let the training set be denoted as $\mathcal{D}=\{ x_{pn},y_{pn} \}, n=1,\ldots,N_p, p=1,\ldots,P$, where $P$ is the number of tasks, and $N_p$ is the number of training samples of the $p$'th task. Here, the $p$'th task is to decide whether or not a sample belongs to the $p$'th category; the sample feature $x_{pn} \in \mathbb{R}^d$ and the corresponding label $y_{pn}=1$ if the $p$'th class label is manually assigned to it, and $0$ otherwise. The general formulation of MTFS is given by
\begin{equation}
\label{eq:MTFS_Formulation}
\mathop{\mathrm{argmin}}_{U} \sum_{p=1}^P \frac{1}{N_p} \sum_{n=1}^{N_p} L(u_p,x_{pn},y_{pn}) + \gamma \sum_{m=1}^d \|u^m\|_2,
\end{equation}
where $U \in \mathbb{R}^{d \times P}$ is a matrix with the model variable $u_p$ for the $p$'th task in a column. The row $u^m$ represents the variables across tasks associated with the $m$'th feature. The regularization is actually the $l_{2,1}$-norm $\|U\|_{2,1}$ which encourages sparsity in rows, and thus the importance of an individual feature is evaluated by simultaneously considering multiple tasks. In this way, different tasks help each other to select features assumed to be shared across tasks. Since the input features for different tasks are commonly the same and only the labels change, the formulation (\ref{eq:MTFS_Formulation}) can be rewritten as
\begin{equation}
\label{eq:MTFS_Reformulation}
\mathop{\mathrm{argmin}}_{U} \sum_{p=1}^P \frac{1}{N} \sum_{n=1}^N L(u_p,x_n,y_{pn}) + \gamma \|U\|_{2,1},
\end{equation}
Following on from this work, Nie et al. \cite{FP-Nie-et-al-NIPS-2010} present a robust formulation by adopting the popular least squares loss for $L$ and utilizing the $l_{2,1}$-norm for both the loss function and regularization, i.e.,
\begin{equation}
\label{eq:RFS_Formulation}
\mathop{\mathrm{argmin}}_{U} \|X^T U - Y\|_{2,1} + \gamma \|U\|_{2,1},
\end{equation}
where $X = [x_1, \ldots, x_N] \in \mathbb{R}^{d \times N}$ is the feature matrix and $Y = [y_1, \ldots, y_N]^T\in \mathbb{R}^{N \times P}$ is the corresponding label matrix. An efficient algorithm is developed in \cite{FP-Nie-et-al-NIPS-2010} to solve it with proven convergence.

\begin{figure*}[!t]
\centering
\includegraphics[width=1.6\columnwidth]{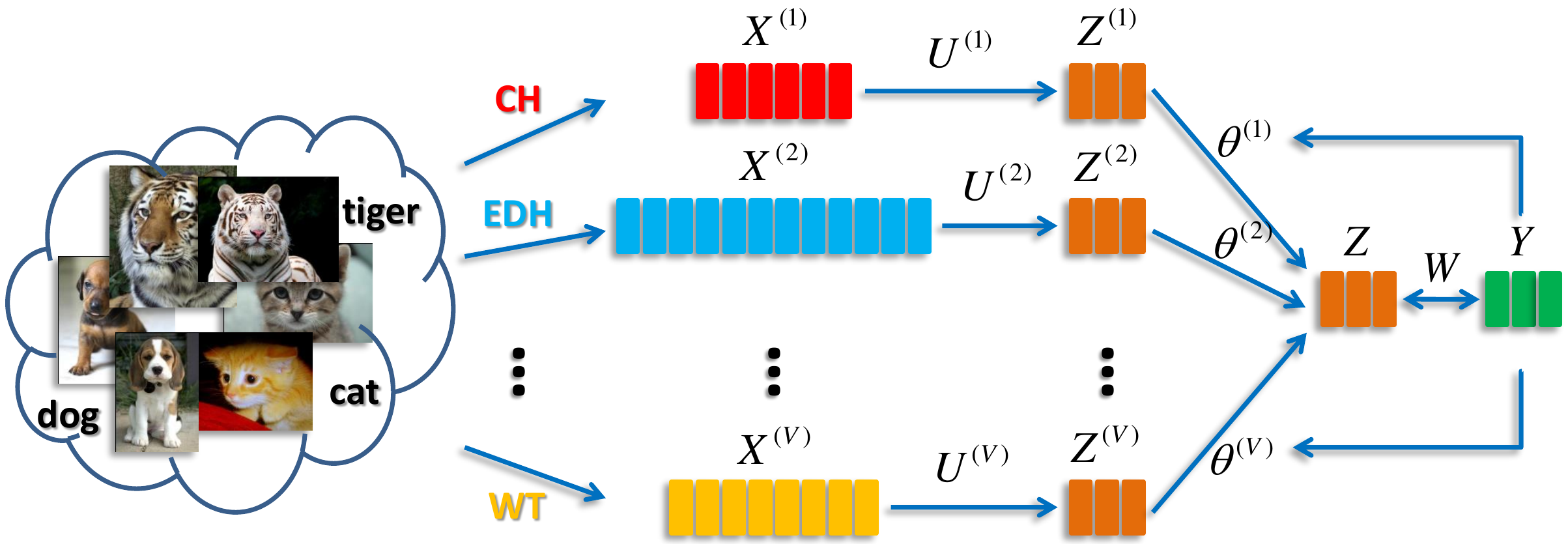}
\caption{The main procedure of the LM3FE framework. Different features, such as CH, EDH, WT, etc., are extracted to represent images from different modalities. The obtained features $X^{(v)},v=1,\ldots,V$ are then transformed into the latent spaces as $\{ Z^{(v)} \}$ using the feature extraction matrices $\{ U^{(v)} \}$. Subsequently, the latent features are weightedly combined as $Z$ to predict the ground-truth label $Y$. Lastly, the extraction matrices $\{ U^{(v)} \}$, the modality combination weights $\{ \theta_v \}$, and the prediction matrix $W$ are learned alternately by minimizing the prediction errors on the training data. The obtained $\{ U^{(v)} \}$ and $\{ \theta_v \}$ are used for final feature extraction.}
\label{fig:Main_Procedure}
\end{figure*}

\section{LM3FE: Large Margin Multi-modal Multi-task Feature Extraction}
\label{sec:LM3FE}

To handle multi-modal image classification, we generalize MTFS and present LM3FE. Fig. \ref{fig:Main_Procedure} illustrates the main procedure of LM3FE. First, various features, such as the color histogram (CH), edge direction histogram (EDH), wavelet texture (WT), etc., are extracted to represent each image in the dataset from different modalities. Then the feature extraction matrix $U^{(v)}$ is learned for each modality by transforming the features from the original space $\mathcal{X}^{(v)}$ into a latent space $\mathcal{Z}^{(v)}$, and the transformed features $Z^{(v)},v=1,\ldots,V$ are weightedly combined as $Z=\sum_{v=1}^V \theta_v Z^{(v)}$. Lastly, the combined latent features are used to predict the class labels $Y$ of the training data. Using an alternating algorithm, the extraction matrices $\{ U^{(v)} \}$, the combination weights $\{ \theta_v \}$, as well as the prediction matrix $W$ are learned by minimizing the prediction errors, and the learned $\{ U^{(v)} \}$ and $\{ \theta_v \}$ are utilized to transform or select features for further classification. The technical details are given below.

\subsection{Problem formulation}
Let $V$ be the number of modalities and $v$ be the modality index. The general formulation of LM3FE is given by
\begin{equation}
\label{eq:LM3FE_General_Formulation}
\begin{split}
\mathop{\mathrm{argmin}}_{W,\{U^{(v)}\},\theta}
& F(W, \{U^{(v)}\}, \theta) = \\
& \Phi(W, \{U^{(v)}\}, \theta) + \Omega(W, \{U^{(v)}\}, \theta), \\
\mathrm{s.t.} \ & \theta_v \geq 0, \ v = 1, \ldots, V,
\end{split}
\end{equation}
where $\Phi(W,\{U^{(v)}\},\theta)$ is the empirical loss given by
\begin{equation}
\notag
\Phi(W,\{U^{(v)}\},\theta) = \sum_{p=1}^P \sum_{n=1}^N g\left(\hbar_p(\{x_n^{(v)}\}_{v=1}^V), y_{pn}\right),
\end{equation}
and $\Omega(W,\{U^{(v)}\},\theta)$ contains all the regularization terms:
\begin{equation}
\notag
\Omega(W,\{U^{(v)}\},\theta) = \gamma_A \|W\|_F^2 + \gamma_B \sum_{v=1}^V \|U^{(v)}\|_{2,1} + \gamma_C \|\theta\|_2^2.
\end{equation}
In problem (\ref{eq:LM3FE_General_Formulation}), $g$ can be any convex loss function such as the least squares loss and hinge loss, and
\begin{equation}
\label{eq:LM3FE_Prediction_Function}
\hbar_p ( \{ x_n^{(v)} \}_{v=1}^V ) = w_p^T \sum_{v=1}^V \theta_v (U^{(v)})^T x_n^{(v)} + b_p,
\end{equation}
is the prediction of the $p$'th category. Here, $w_p$ is the $p$'th column of the large margin prediction matrix $W$, $b_p$ is the bias and can be absorbed into the learning of $w_p$, $U^{(v)}$ is the feature selection matrix for the $v$'th modality, and $\theta = [\theta_1, \ldots , \theta_V]^T$ is a vector of the modality integration weights. The regularization term $\|W\|_F^2$ is used to control model complexity, $\| U^{(v)} \|_{2,1}$ is to perform multi-task feature selection, and $\|\theta\|_2^2$ is served as a smoothing term to avoid model over-fitting to only one or a small number of views. All $\gamma_A$, $\gamma_B$ and $\gamma_C$ are positive trade-off parameters. The features of different modalities $x_n^{(v)}, v = 1,\ldots,V$ are assumed to have been normalized.

In this paper, we choose $g$ as the hinge loss so that the extracted features are particularly suitable for classification, i.e., $g(\hbar_p(\{x_n^{(v)}\}_{v=1}^V), y_{pn}) = \mathrm{max}(0, 1-y_{pn}\hbar_p(\{x_n^{(v)}\}_{v=1}^V))$. The loss function $g$ is non-differentiable. Hence, we first smooth the loss according to \cite{Y-Nesterov-MP-2005}, and the smoothed version of the hinge loss $g$ can be given by
\begin{equation}
\label{eq:Smoothed_g}
g^\sigma = \mathop{\mathrm{max}}_{\nu_p \in \mathcal{Q}} \nu_{pn}\left(1 - y_{pn}\hbar_p(\{x_n^{(v)}\})\right) - \frac{\sigma}{2} \|x_n\|_\infty \nu_{pn}^2,
\end{equation}
where $\mathcal{Q} = \{ \nu_{pn}: 0 \leq \nu_{pn} \leq 1, \nu_{pn} \in \mathbb{R}^N \}$, $\nu_p = [\nu_{p1}, \nu_{p2}, \ldots , \nu_{pN}]^T$, $x_n$ is a concatenation of $\{ x_n^{(v)} \}$ and $\sigma$ is the smooth parameter, which is set at $5$ in this paper. By setting the objective function of (\ref{eq:Smoothed_g}) to zero and then projecting $\nu_{pn}$ on $\mathcal{Q}$, we obtain the following solution:
\begin{equation}
\label{eq:Form_nu}
\nu_{pn} = \mathrm{median} \left\{ \frac{1 - y_{pn}\hbar_p(\{x_n^{(v)}\})}{\sigma \|x_n\|_\infty}, 0, 1 \right\}.
\end{equation}
By substituting the solution (\ref{eq:Form_nu}) back into (\ref{eq:Smoothed_g}), we have the piece-wise approximation of $g$, i.e.,
\begin{equation}
\label{eq:PieceWise_g}
\begin{split}
& g^\sigma = \\
& \left\{
\begin{array}{cc}
0, & \ y_{pn}\hbar_p(\{x_n^{(v)}\}) > 1; \\
\left(1 - y_{pn}\hbar_p(\{x_n^{(v)}\})\right) - \frac{\sigma}{2} \|x_n\|_\infty, & \ \begin{array}{c}
                                                                                        y_{pn}\hbar_p(\{x_n^{(v)}\}) < \\
                                                                                        1 - \sigma \|x_n\|_\infty;
                                                                                     \end{array} \\
\frac{\left( 1 - y_{pn}\hbar_p(\{x_n^{(v)}\}) \right)^2}{2 \sigma \|x_n\|_\infty}, & \ \mathrm{otherwise}.
\end{array}
\right.
\end{split}
\end{equation}
%\begin{numcases}
%{}{}
%\begin{split}
%\label{eq:PieceWise_g}
%0, \ \ \ \ \ \ \ \ \ \ \ \ \ \ \ \ \ & \ y_{pn}\hbar_p(\{x_n^{(v)}\}) > 1; \\
%\left(1 - y_{pn}\hbar_p(\{x_n^{(v)}\})\right) - \frac{\sigma}{2} \|x_n\|_\infty, & \ \begin{array}{c}
%                                                                                        y_{pn}\hbar_p(\{x_n^{(v)}\}) < \\
%                                                                                        1 - \sigma \|x_n\|_\infty;
%                                                                                     \end{array} \\
%\frac{\left( 1 - y_{pn}\hbar_p(\{x_n^{(v)}\}) \right)^2}{2 \sigma \|x_n\|_\infty}, \ \ \ \ \ \ & \ \mathrm{otherwise}.
%\end{split}
%\end{numcases}
Using the smoothed hinge loss $g^\sigma$ in problem (\ref{eq:LM3FE_General_Formulation}), we can obtain the solutions for the three sets of variables $W$, $\{ U^{(v)} \}$, and $\theta$ by optimizing them alternately until convergence.

\subsection{Optimization}
Due to the efficiency of Nesterov's optimal gradient method \cite{Y-Nesterov-MP-2005}, we adopt it to solve the sub-problems w.r.t. $W$ and $\{ U^{(v)} \}$. The sub-problem w.r.t. $\theta$ is a constrained optimization problem and thus is solved using a non-negative optimal gradient method \cite{NY-Guan-et-al-TSP-2012}. The details of optimizing each sub-problem are presented as follows:

\subsubsection{Update for $W$}
By using the smoothed hinge loss $g^\sigma$, and fixing $\{ U^{(v)} \}$ and $\theta$, the problem (\ref{eq:LM3FE_General_Formulation}) becomes:
\begin{equation}
\label{eq:Formulation_wrt_W}
\begin{split}
\mathop{\mathrm{argmin}}_W &\ F(W) = \sum_{p=1}^P \sum_{n=1}^N g^\sigma\left( \hbar_p(\{x_n^{(v)}\}), y_{pn} \right) + \gamma_A \|W\|_F^2, \\
& = \sum_{p=1}^P \left( \sum_{n=1}^N g^\sigma\left(\hbar_p(\{x_n^{(v)}\}), y_{pn}\right) + \gamma_A \|w_p\|_2^2 \right),
\end{split}
\end{equation}
The problem (\ref{eq:Formulation_wrt_W}) can be decomposed to solving for each $w_p$ independently, and the formulation of the $p$'th sub-problem is given by
\begin{equation}
\label{eq:Formulation_wrt_wp}
\mathop{\mathrm{argmin}}_{w_p} F(w_p) = \Phi(w_p) + \Omega(w_p),
\end{equation}
where $\Phi(w_p) = \sum_{n=1}^N g^\sigma(z_n,y_{pn},w_p)$ and $\Omega(w_p) = \gamma_A \|w_p\|_2^2$. Here, $g^\sigma(z_n,y_{pn},w_p)$ is a reformulation of $g^\sigma$ w.r.t. $w_p$. That is, the term $\hbar_p(\{x_n^{(v)}\})$ in $g^\sigma$ is rewritten as $\hbar_p(z_n,w_p) = w_p^T z_n + b_p$, where $z_n = \sum_{v=1}^V \theta_v (U^{(v)})^T x_n^{(v)}$. We adopt Nesterov's method to solve problem (\ref{eq:Formulation_wrt_wp}) since it is able to achieve the optimal convergence rate at $O(1/t^2)$, where $t$ is the number of iterations. To utilize Nesterov's method for optimization, we have to compute the gradient of the smoothed hinge loss to determine the direction of the descent, as well as the Lipschitz constant to determine the step size of each iteration. We summarize the results in the following theorem.
\begin{thm}
\label{thm:Gradient_Lipschitz_Cst_g_wp}
The sum of the gradient of the smoothed hinge loss $g^\sigma$ w.r.t. $w_p$ over all the $N$ samples is
\begin{equation}
\label{eq:Gradient_g_wp}
\frac{\partial g^\sigma(w_p)}{\partial w_p} = -Z Y_p \nu_p,
\end{equation}
where $Z = [z_1,z_2, \ldots ,z_N]$, $Y_p = \mathrm{diag}(y_p)$ and $y_p=[y_{p1},y_{p2}, \ldots ,y_{pN}]^T$. The Lipschitz constant of $g^\sigma(w_p)$ is
\begin{equation}
\label{eq:Lipschitz_Cst_g_wp}
L_g^\sigma(w_p) = \frac{N}{\sigma} \mathop{\mathrm{max}}_n \frac{\|z_n z_n^T\|_2}{\|x_n\|_\infty}.
\end{equation}
\end{thm}
The proof can be found in the Appendix.

In addition, it is easy to deduce that the gradient of $\Omega(w_p)$ is $2 \gamma_A w_p$ and the Lipschitz constant is $2\gamma_A$. Therefore, the gradient of $F(w_p)$ is
\begin{equation}
\label{eq:Gradient_F_wp}
\frac{\partial F(w_p)}{\partial w_p} = -Z Y_p \nu_p + 2 \gamma_A w_p,
\end{equation}
and the Lipschitz constant is
\begin{equation}
\label{eq:Lipschitz_Cst_F_wp}
L_F(w_p) = \frac{N}{\sigma} \mathop{\mathrm{max}}_n \frac{\|z_n z_n^T\|_2}{\|x_n\|_\infty} + 2 \gamma_A.
\end{equation}
Based on the obtained gradient and Lipschitz constant, we apply Nesterov's method to minimize the smoothed primal $F(w_p)$. In the $t$'th iteration round, two auxiliary optimizations are constructed and their solutions are used to build the solution to problem (\ref{eq:Formulation_wrt_wp}). We use $w_p^t$, $y^t$ and $z^t$ to represent the solutions of (\ref{eq:Formulation_wrt_wp}) and its two auxiliary optimizations at the $t$'th iteration round, respectively. The Lipschitz constant of $F(w_p)$ is $L_F(w_p)$ and the two auxiliary optimizations are
\begin{equation}
\notag
\begin{split}
\mathop{\mathrm{min}}_y & \langle \nabla F(w_p^t), y-w_p^t \rangle + \frac{L_F(w_p)}{2} \|y - w_p^t\|_2^2, \\
\mathop{\mathrm{min}}_z & \sum_{i=0}^t \frac{i+1}{2} [F(w_p^i) + \langle \nabla F(w_p^i), z-w_p^i \rangle] \\
& + \frac{L_F(w_p)}{2} \|z - \hat{w}_p\|_2^2,
\end{split}
\end{equation}
where $\hat{w}_p$ is an estimated solution of $w_p$. By directly setting the gradients of the two objective functions in the auxiliary optimizations as zeros, we can obtain $y^t$ and $z^t$, respectively,
\begin{eqnarray}
\label{eq:Form_yt_wp}
&& y^t = w_p^t - \frac{1}{L_F(w_p)} \nabla F(w_p^t), \\
\label{eq:Form_zt_wp}
&& z^t = \hat{w}_p - \frac{1}{L_F(w_p)} \sum_{i=0}^t \frac{i+1}{2} \nabla F(w_p^i).
\end{eqnarray}
The solution after the $t$'th iteration round is the weighted sum of $y^t$ and $z^t$, i.e.,
\begin{equation}
\label{eq:Form_wp}
w_p^{t+1} = \frac{2}{t+3} z^t + \frac{t+1}{t+3} y^t.
\end{equation}
The stopping criterion is $|F(w_p^{t+1})-F(w_p^t)| / |F(w_p^{t+1})-F(w_p^0)| < \epsilon$, where $\epsilon$ is a predefined threshold which we set as $10^{-3}$ in this paper. The initialization $w_p^0$ and the estimated solution $\hat{w}_p$ are set as the zero vectors. The bias $b_p$ can be learned by letting $w_p = [w_p; b_p]$ and $Z = [Z; e^T]$, where $e$ is a vector of all ones. The gradient of $\Omega(\cdot)$ w.r.t. $b_p$ is zero since it is not penalized using a $l_2$-norm. The last entry of the output solution $w_p$ is the bias $b_p$.

\subsubsection{Update for $\{ U^{(v)} \}$}
For fixed $W$ and $\theta$, and by using the smoothed hinge loss $g^\sigma$, the problem (\ref{eq:LM3FE_General_Formulation}) becomes
\begin{equation}
\label{eq:Formulation_wrt_Us}
\begin{split}
\mathop{\mathrm{argmin}}_{\{U^{(v)}\}} & \ F(\{U^{(v)}\}) \\
= & \sum_{p=1}^P \sum_{n=1}^N g^\sigma\left( \hbar_p(\{x_n^{(v)}\}), y_{pn} \right) + \gamma_B \sum_{v=1}^V \|U^{(v)}\|_{2,1},
\end{split}
\end{equation}
We can update for each $U^{(v)}$ alternately by fixing all the other $U^{(v')}, v' \neq v$ until convergence. The sub-problem for optimizing each $U^{(v)}$ is given by
\begin{equation}
\label{eq:Formulation_wrt_Uv}
\mathop{\mathrm{argmin}}_{U^{(v)}} \ F(U^{(v)}) = \Phi(U^{(v)}) + \Omega(U^{(v)}),
\end{equation}
where $\Phi(U^{(v)}) = \sum_{p=1}^P \sum_{n=1}^N g^\sigma(x_n^{(v)}, y_{pn}, U^{(v)})$, and $\Omega(U^{(v)}) = \gamma_B \|U^{(v)}\|_{2,1}$. Here, $g^\sigma(x_n^{(v)},y_{pn},U^{(v)})$ is a reformulation of $g^\sigma$ w.r.t. $U^{(v)}$. That is, the term $\hbar_p(\{ x_n^{(v)} \})$ in $g^\sigma$ is rewritten as $\hbar_p(x_n^{(v)}, U^{(v)}) = \theta_v w_p^T (U^{(v)})^T x_n^{(v)} + c_{pn}^{(v)}$, where $c_{pn}^{(v)} = w_p^T \sum_{v' \neq v} \theta_v' (U^{(v')})^T x_n^{(v')} + b_p$ consists of all the terms that are irrelevant to $U^{(v)}$. Similar to the optimization of $w_p$, we adopt Nesterov's method to solve problem (\ref{eq:Formulation_wrt_Uv}), and the gradient and Lipschitz constant are summarized in the following theorem.
\begin{thm}
\label{thm:Gradient_Lipschitz_Cst_Uv}
The sum of the gradient of the smoothed hinge loss $g^\sigma$ w.r.t. $U^{(v)}$ over all the $N$ samples and $P$ class labels is
\begin{equation}
\label{eq:Gradient_g_Uv}
\frac{\partial g^\sigma(U^{(v)})}{\partial U^{(v)}} = \sum_{p=1}^P -\theta_v X^{(v)} Y_p \nu_p w_p^T,
\end{equation}
The Lipschitz constant of $g^\sigma(U^{(v)})$ is
\begin{equation}
\label{eq:Lipschitz_Cst_g_Uv}
L_g^\sigma(U^{(v)}) = \frac{PN\theta_v^2}{\sigma} \mathop{\mathrm{max}}_p \mathop{\mathrm{max}}_n \frac{\|x_n^{(v)} w_p^T\|_2 \|(x_n^{(v)})^T\|_2 \|w_p\|_2}{\|x_n\|_\infty}.
\end{equation}
\end{thm}
The proof can be found in the Appendix.

In addition, it is well known that the gradient of $\Omega(U^{(v)})$ is $2 \gamma_B D^{(v)} U^{(v)}$ \cite{FP-Nie-et-al-NIPS-2010}, and the Lipschitz constant is $2 \gamma_B \|D^{(v)}\|_2$, where $D^{(v)}$ is a diagonal matrix with the entry $D_{ii}^{(v)} = \frac{1}{2\|u^{(v),i}\|_2}$, and $u^{(v),i}$ is the $i$'th row of $D^{(v)}$. Note that although $D^{(v)}$ is dependent on $U^{(v)}$, we regard it as a constant term in the calculation of the Lipschitz constant and update it after each iteration of Nesterov's method. A modified Nesterov method for solving $U^{(v)}$ and its convergence analysis will be presented later. The gradient of $F(U^{(v)})$ is then
\begin{equation}
\label{eq:Gradient_F_Uv}
\frac{\partial F(U^{(v)})}{\partial U^{(v)}} = \sum_{p=1}^P -\theta_v X^{(v)} Y_p \nu_p w_p^T + 2 \gamma_B D^{(v)} U^{(v)},
\end{equation}
and the Lipschitz constant is
\begin{equation}
\label{eq:Lipschitz_Cst_F_Uv}
\begin{split}
& L_F(U^{(v)}) = \\
& \frac{PN\theta_v^2}{\sigma} \mathop{\mathrm{max}}_p \mathop{\mathrm{max}}_n \frac{\|x_n^{(v)} w_p^T\|_2 \|(x_n^{(v)})^T\|_2 \|w_p\|_2}{\|x_n\|_\infty} + 2 \gamma_B \|D^{(v)}\|_2,
\end{split}
\end{equation}
Based on the obtained gradient and Lipschitz constant, we apply Nesterov's method to minimize the smoothed primal $F(U^{(v)})$. We use $U_t^{(v)}$, $Y^t$ and $Z^t$ to represent the solutions of (\ref{eq:Formulation_wrt_Uv}) and its two auxiliary optimizations at the $t$'th iteration round, respectively. The Lipschitz constant of $F(U^{(v)})$ is $L_F(U_t^{(v)})$, which changes at each iteration, and the two auxiliary optimizations are
\begin{equation}
\notag
\begin{split}
\mathop{\mathrm{min}}_Y & \langle \nabla F(U_t^{(v)}), Y-U_t^{(v)} \rangle + \frac{L_F(U_t^{(v)})}{2} \|Y - U_t^{(v)}\|_F^2, \\
\mathop{\mathrm{min}}_Z & \sum_{i=0}^t \frac{i+1}{2} [F(U_i^{(v)}) + \langle \nabla F(U_i^{(v)}), Z-U_i^{(v)} \rangle] \\
& + \frac{L_F(U_t^{(v)})}{2} \|Z - \hat{U}^{(v)}\|_F^2,
\end{split}
\end{equation}
where $\hat{U}^{(v)}$ is an estimated solution of $U^{(v)}$. By directly setting the gradients of the two objective functions in the auxiliary optimizations as zeros, we can obtain $Y^t$ and $Z^t$, respectively,
\begin{eqnarray}
\label{eq:Form_Yt_Uv}
&& Y^t = U_t^{(v)} - \frac{1}{L_F(U_t^{(v)})} \nabla F(U_t^{(v)}), \\
\label{eq:Form_Zt_Uv}
&& Z^t = \hat{U}^{(v)} - \frac{1}{L_F(U_t^{(v)})} \sum_{i=0}^t \frac{i+1}{2} \nabla F(U_i^{(v)}).
\end{eqnarray}
The solution after the $t$'th iteration round is the weighted sum of $Y^t$ and $Z^t$, i.e.,
\begin{equation}
\label{eq:Form_Uv}
U_{t+1}^{(v)} = \frac{2}{t+3} Z^t + \frac{t+1}{t+3} Y^t.
\end{equation}
The diagonal matrix $D^{(v)}$ is then updated using the obtained $U_{t+1}^{(v)}$, and we summarize the algorithm for solving $U^{(v)}$ in Algorithm \ref{alg:Optimization_Uv}. The stopping criterion is $|F(U_{t+1}^{(v)})-F(U_t^{(v)})| / |F(U_{t+1}^{(v)})-F(U_0^{(v)})| < \epsilon$. The initialization $U_0^{(v)}$ and the estimated solution $\hat{U}^{(v)}$ are set as the results of the previous iterations in the alternating optimization of ${U^{(v)}}$. The convergence of Algorithm \ref{alg:Optimization_Uv} is guaranteed by the following theorem.
\begin{thm}
\label{thm:Convergence_Opt_Uv}
The objective of the problem (\ref{eq:Formulation_wrt_Uv}) will monotonically decrease in each iteration of Algorithm \ref{alg:Optimization_Uv}.
\end{thm}
The proof can be found in the Appendix.

\begin{algorithm}[!t]%[htb]
\caption{Modified Nesterov optimal gradient method for solving $U^{(v)}$.}
\label{alg:Optimization_Uv}
\begin{algorithmic}[1]
\renewcommand{\algorithmicrequire}{\textbf{Input:}}
\REQUIRE $\{ X^{(v)} \}$, $\{ Y_p \}$, $W$, $\theta$ and $U^{(v')}, v' \neq v$.
\renewcommand{\algorithmicrequire}{\textbf{Parameters:}}
\REQUIRE $\gamma_B$.
\renewcommand{\algorithmicensure}{\textbf{Output:}}
\ENSURE $U^{(v)}$.
\STATE{Initialize $U_t^{(v)}$ as the results of the previous iterations in the alternating optimization of $\{ U^{(v)} \}$, and $D_t^{(v)}$ is calculated based on $U_t^{(v)}$. Set $t=0$.}
\renewcommand{\algorithmicrepeat}{\textbf{Repeat}}
\renewcommand{\algorithmicuntil}{\textbf{Until convergence}}
\REPEAT
\STATE{Calculate the gradient and Lipschitz constant using (\ref{eq:Gradient_F_Uv}) and (\ref{eq:Lipschitz_Cst_F_Uv});}
\STATE{Calculate $U_{t+1}^{(v)}$ using (\ref{eq:Form_Yt_Uv}), (\ref{eq:Form_Zt_Uv}), and (\ref{eq:Form_Uv});}
\STATE{Update the diagonal matrix $D_{t+1}^{(v)}$ based on $U_{t+1}^{(v)}$;}
\STATE{$t \leftarrow t+1$.}
\UNTIL
\end{algorithmic}
\end{algorithm}

\subsubsection{Update for $\{\theta_v\}$}
For fixed $W$ and $\{ U^{(v)} \}$, and by using the smoothed hinge loss $g^\sigma$, the problem (\ref{eq:LM3FE_General_Formulation}) becomes
\begin{equation}
\label{eq:Formulation_wrt_theta}
\begin{split}
\mathop{\mathrm{argmin}}_{\theta} \ & F(\theta) = \Phi(\theta) + \Omega(\theta), \\
\mathrm{s.t.} \ & \theta_v \geq 0, \ v = 1, \ldots, V,
\end{split}
\end{equation}
where $\Phi(\theta) = \sum_{p=1}^P \sum_{n=1}^N g^\sigma(z_{pn}, y_{pn}, \theta)$, and $\Omega(\theta) = \gamma_C \|\theta\|_2^2$. Here, $g^\sigma(z_{pn}, y_{pn}, \theta)$ is a reformulation of $g^\sigma$ w.r.t. $\theta$. That is, the term $\hbar_p(\{ x_n^{(v)} \})$ in $g^\sigma$ is rewritten as $\hbar_p(z_{pn}, \theta) = \theta^T z_{pn} + b_p$, where $z_{pn} = [z_{pn}^{(1)}, z_{pn}^{(2)}, \ldots, z_{pn}^{(V)}]^T$ with each $z_{pn}^{(v)} = w_p^T (U^{(v)})^T x_n^{(v)}$. In contrast to the optimization of $w_p$ and $U^{(v)}$, there is an additional non-negative constraint on $\theta$. We thus utilize the non-negative optimal gradient method (OGM) \cite{NY-Guan-et-al-TSP-2012} to solve (\ref{eq:Formulation_wrt_theta}). Similar to the gradient of $g^\sigma$ w.r.t. $w_p$, the gradient of $g^\sigma$ w.r.t. $\theta$ for the $n$'th sample is
\begin{equation}
\label{eq:Gradient_g_theta_xn}
\frac{\partial g^\sigma(z_{pn}, y_{pn}, \theta)}{\partial \theta} = - y_{pn} z_{pn} \nu_{pn},
\end{equation}
Thus the sum of the gradient over all the $N$ samples and $P$ labels is
\begin{equation}
\label{eq:Gradient_g_theta}
\frac{\partial g^\sigma(\theta)}{\partial \theta} = \frac{\partial \sum_{p=1}^P \sum_{n=1}^N g^\sigma(z_{pn}, y_{pn}, \theta)}{\partial \theta} = \sum_p (-Z_p Y_p \nu_p),
\end{equation}
where $Z_p = [z_{p1},z_{p2}, \ldots ,z_{pN}]$. The Lipschitz constant of $g^\sigma(\theta)$ is
\begin{equation}
\label{eq:Lipschitz_Cst_g_theta}
L_g^\sigma(\theta) = \frac{PN}{\sigma} \mathop{\mathrm{max}}_p \mathop{\mathrm{max}}_n \frac{\|z_{pn} z_{pn}^T\|_2}{\|x_n\|_\infty}.
\end{equation}
In addition, the gradient of $\Omega(\theta)$ is $2 \gamma_C \theta$ and the Lipschitz constant is $2 \gamma_C$. Therefore, the gradient of $F(\theta)$, is
\begin{equation}
\label{eq:Gradient_F_theta}
\frac{\partial F(\theta)}{\partial \theta} = \sum_{p=1}^P (-Z_p Y_p \nu_p) + 2 \gamma_C \theta,
\end{equation}
and the Lipschitz constant is
\begin{equation}
\label{eq:Lipschitz_Cst_F_theta}
L_F(\theta) = \frac{PN}{\sigma} \mathop{\mathrm{max}}_p \mathop{\mathrm{max}}_n \frac{\|z_{pn} z_{pn}^T\|_2}{\|x_n\|_\infty} + 2 \gamma_C.
\end{equation}
Based on the obtained gradient and Lipschitz constant, we apply the optimal gradient method presented in \cite{NY-Guan-et-al-TSP-2012} to minimize the smoothed $F(\theta)$. We summarize the procedure in Algorithm \ref{alg:Optimization_theta} cited from \cite{NY-Guan-et-al-TSP-2012} (Algorithm 1 therein), where the operator $P[x]$ projects all the negative entries of $x$ to zero. The algorithm is guaranteed to converge and achieves the optimal convergence rate $O(1/t^2)$ according to \cite{NY-Guan-et-al-TSP-2012}. The stopping criterion we utilized here is $|F(\theta^{t+1})-F(\theta^t)| / |F(\theta^{t+1})-F(\theta^0)| < \epsilon$. The initialization $\theta^0$ is set as the result of the previous iterations in the alternating optimization of $W$, $\{ U^{(v)} \}$, and $\theta$.

\begin{algorithm}[!t]%[htb]
\caption{The optimal gradient method for solving $\theta$.}
\label{alg:Optimization_theta}
\begin{algorithmic}[1]
\renewcommand{\algorithmicrequire}{\textbf{Input:}}
\REQUIRE $\{ X^{(v)} \}$, $\{ Y_p \}$, $W$, and $\{ U^{(v)} \}$.
\renewcommand{\algorithmicrequire}{\textbf{Parameters:}}
\REQUIRE $\gamma_C$.
\renewcommand{\algorithmicensure}{\textbf{Output:}}
\ENSURE $\theta$.
\STATE{Initialize $\theta^0$ as the results of the previous iterations in the alternating optimization of $W$, $\{ U^{(v)} \}$, and $\theta$. Set $\rho_0=1$, $y^0=\theta^0$ and $t=0$.}
\renewcommand{\algorithmicrepeat}{\textbf{Iterate}}
\renewcommand{\algorithmicuntil}{\textbf{Until convergence}}
\REPEAT
\STATE{Update $\theta^{t+1} = P[y^t - \frac{1}{L_F(\theta)} \nabla F(\theta^t)]$;}
\STATE{Update $\rho_{t+1} = (1 + \sqrt{4\rho_t^2+1})/2$;}
\STATE{Update $y^{t+1} = \theta^{t+1} + \frac{\rho_t-1}{\rho_{t+1}} (\theta^{t+1}-\theta^t)$;}
\STATE{$t \leftarrow t+1$.}
\UNTIL
\end{algorithmic}
\end{algorithm}

We now summarize the learning procedure of the proposed LM3FE in Algorithm \ref{alg:Optimization_LM3FE}. The stopping criterion for terminating the algorithm is the difference of the objective value $F(W, \{ U^{(v)} \}, \theta)$ between two consecutive steps. Alternatively, we can stop the iterations when the variation in $\theta$ is smaller than a predefined threshold. Our implementation is based on the difference of the objective value, i.e., if $|O_{k+1}-O_k| / |O_{k+1}-O_0| < \epsilon$, then the iteration stops, where $O_k$ is the objective value of the $k$'th iteration step. Once the solutions of $\{ U^{(v)} \}$ and $\{ \theta_v \}$ have been obtained, we can use them in two different ways: 1) feature selection, i.e., sort the features according to $\| u^{(v), i} \|_2,i=1,\ldots,d_v$ in descending order for the $v$'th modality and then train additional classifiers on top-ranking features; 2) feature transformation, i.e., use $\sum_{v=1}^V \theta_v (U^{(v)})^T x_n^{(v)}$ as the projected features in the low-dimensional space and then train additional classifiers on these low-dimensional features. Both of these strategies are investigated in the experiments.

\begin{algorithm}[!t]%[htb]
\caption{The optimization procedure of the proposed LM3FE algorithm.}
\label{alg:Optimization_LM3FE}
\begin{algorithmic}[1]
\renewcommand{\algorithmicrequire}{\textbf{Input:}}
\REQUIRE The feature matrices of different modalities $\{ X^{(v)} \}$ and corresponding label matrix $Y$.
\renewcommand{\algorithmicrequire}{\textbf{Parameters:}}
\REQUIRE $\sigma$, $\gamma_A$, $\gamma_B$, and $\gamma_C$.
\renewcommand{\algorithmicensure}{\textbf{Output:}}
\ENSURE $W$, $\{ U^{(v)} \}$, and $\theta$.
\STATE{Initialize $U^{(v)}, v = 1,\ldots,V$ as random matrices, and $\theta_v=1/V$. Set $\sigma=5$.}
\renewcommand{\algorithmicrepeat}{\textbf{Iterate}}
\renewcommand{\algorithmicuntil}{\textbf{Until convergence}}
\REPEAT
\STATE{Solve for each $w_p, p \in \{1,\ldots,P\}$ independently by optimizing (\ref{eq:Formulation_wrt_wp}) using the Nestrov optimal gradient method;}
\STATE{Alternately solve for each $U^{(v)}, v \in \{1,\ldots,V\}$ by optimizing (\ref{eq:Formulation_wrt_Uv}) using the Algorithm \ref{alg:Optimization_Uv} until converges;}
\STATE{Solve for $\theta$ by optimizing (\ref{eq:Formulation_wrt_theta}) using the Algorithm \ref{alg:Optimization_theta}.}
\UNTIL
\end{algorithmic}
\end{algorithm}

\subsection{Convergence analysis}

In this section, we discuss the convergence of the proposed LM3FE algorithm. Let the initialized value of the objective function (\ref{eq:LM3FE_General_Formulation}) be $F(W^K, \{ U_K^{(v)} \}, \theta^K)$. Since (\ref{eq:Formulation_wrt_W}) is convex, we have $F(W^{K+1}, \{ U_K^{(v)}\}, \theta^K ) \leq F(W^K, \{ U_K^{(v)} \}, \theta^K )$. The problem (\ref{eq:Formulation_wrt_Us}) is solved alternately for each $U^{(v)}$ and all the other $U^{(v')}, v' \neq v$ are fixed. Theorem \ref{thm:Convergence_Opt_Uv} ensures that the objective value of (\ref{eq:Formulation_wrt_Us}) decreases at each iteration of the alternating procedure, i.e., $F(U_{k+1}^{(v)}, \{ U_k^{(v')} \}_{v' \neq v}) \leq F(\{ U_k^{(v)} \})$. This indicates that $F(W^{K+1}, \{ U_{K+1}^{(v)} \}, \theta^K) \leq F(W^{K+1}, \{ U_K^{(v)} \}, \theta^K)$. Finally, we have $F(W^{K+1}, \{ U_{K+1}^{(v)} \}, \theta^{K+1}) \leq F(W^{K+1}, \{ U_{K+1}^{(v)} \}, \theta^K )$ according to the convergence result provided in \cite{NY-Guan-et-al-TSP-2012}. Therefore, the convergence of our algorithm is guaranteed.

\subsection{Complexity analysis}

To analyze the time complexity of the proposed LM3FE algorithm, we first present the computational cost of optimizing $W$, $\{ U^{(v)} \}$, and $\theta$ respectively: 1) optimizing $W$ includes a pre-calculation of the $P \times N$ matrix $Z$, and the optimizations of $P$ sub-problems, where the $p$'th optimization is to find the solution for $w_p$. The time cost of the pre-calculation is $O(\bar{d}_v VPN)$, where $\bar{d}_v$ is the average feature dimensions. In each iteration of solving for $w_p$, the main cost is spent on the calculation of the gradient $\nabla F(w_p^t)$, where the complexity is given by $O(P N^2)$. Therefore, the complexity of optimizing $W$ is $O(\bar{d}_v VPN) + k_1 O(P^2 N^2)$, where $k_1$ is the number of iterations and is typically small due to the fast convergence property of Nesterov's method; 2) the solutions for $\{ U^{(v)} \}$ are obtained by alternating between each $U^{(v)}$. Although the Lipschitz constant $L_F(U^{(v)})$ changes at each iteration in the optimization of $U^{(v)}$, the first term is a constant value and can be calculated beforehand in $O(\bar{d}_v V P^3 N)$ time. In each iteration of optimizing $U^{(v)}$, the complexity is dominated by the time cost $O(\bar{d}_v (P^2 N + P N^2))$ of computing the gradient $\nabla F(U_t^{(v)})$. Therefore, the complexity of optimizing $\{ U^{(v)} \}$ is $O(\bar{d}_v V P^3 N) + k_2^{out} V k_2^{in} O(\bar{d}_v (P^2 N + P N^2))$, where $k_2^{in}$ is the number of iterations of the modified Nesterov method presented in Algorithm \ref{alg:Optimization_Uv}, and it is often small. Here, $k_2^{out}$ is the number of iterations of the alternating procedure for updating $\{ U^{(v)} \}$, and we set it as $1$ since this setting does not impact on the convergence property of the proposed LM3FE algorithm, and we found little decline in performance in our experiments; 3) the optimization of $\theta$ involves a pre-calculation of the matrices $\{ Z_p \}$, the time cost of which is $O(\bar{d}_v VPN)$. The complexity of optimizing $\theta$ is dominated by the time cost $O(V P N^2)$ of computing the gradient $\nabla F(\theta^t)$, and thus is given by $O(\bar{d}_v V P N) + k_3 O(V P N^2)$, where $k_3$ is the number of iterations of the optimal gradient method presented in Algorithm \ref{alg:Optimization_theta} and is also typically small.

Lastly, according to the above analysis, we conclude that the computational cost of LM3FE is $K( O(\bar{d}_v V P^3 N) + k_1 O(P^2 N^2 ) + V k_2^{in} O(\bar{d}_v (P^2 N + P N^2)) + k_3 O(V P N^2) )$, where $K$ is the number of iterations in Algorithm \ref{alg:Optimization_LM3FE}. The second and last terms are usually very small compared with the third term, and thus the time complexity is approximately $K( O(\bar{d}_v V P^3 N) + V k_2^{in} O(\bar{d}_v (P^2 N + P N^2)) )$. This is linear w.r.t. the average feature dimension, and quadratic in the number of training samples, and thus is not very high.

\section{Experiments}
\label{sec:Experiments}

In the experiments, we evaluate the effectiveness of the proposed LM3FE algorithm on the image classification problem by first applying it for feature selection and then regarding it as a feature transformation approach. Prior to these evaluations, we present the used datasets and features, as well as our experimental settings.

\subsection{Datasets, features and evaluation criteria}

Our experiments are conducted on two challenging real-world image datasets, NUS-WIDE (NUS) \cite{TS-Chua-et-al-CIVR-2009} and MIR Flickr (MIR) \cite{MJ-Huiskes-and-MS-Lew-ICMR-2008}.

The NUS dataset contains $269,648$ images, and our experiments are conducted on a subset that consists of $16,519$ images belonging to $12$ animal concepts: bear, bird, cat, cow, dog, elk, fish, fox, horse, tiger, whale, and zebra. We randomly split the images into a training set of $8,263$ images and a test set of $8,256$ images. Distinguishing between these concepts is very challenging, since many of them are similar to one another, e.g., cat and tiger. We randomly choose $\{ 4,6,8 \}$ labeled instances for each concept in the training set to determine the performance of the compared methods w.r.t. the number of labeled instances. Six different features, namely $500$-D bag of visual words based on SIFT \cite{DG-Lowe-IJCV-2004} descriptors, $64$-D color histogram, $144$-D color auto-correlogram, $73$-D edge direction histogram, $128$-D wavelet texture, and $225$-D block-wise color moments are used to represent each image. The $1$-nearest neighbour ($1$NN) classifier is adopted and the evaluation criterion is classification accuracy. We also use F1-score (\cite{M-Sokolova-and-G-Lapalme-JIPM-2009}) to measure the performance for individual classes, and macroF1 score, which is an average of the F1-scores for all classes is served as another criterion to evaluate the performance of different methods.

The MIR dataset consists of $38$ classes and $25,000$ images, which are randomly split into equally sized training and test sets. The number of labeled positive instances are set as $\{ 20,30,50 \}$ for each category, and the same number of negative samples is selected. The images are represented by three features: $1000$-D bag of visual words based on the local SIFT, $512$-D global GIST descriptions \cite{A-Oliva-and-A-Torralba-IJCV-2001}, and the tags ($457$-D). MIR is a multi-label dataset (i.e., an image can have multiple labels). Therefore, we adopt regularized least square (RLS) as the classifier and introduce a popular criterion in multi-label classification for evaluation, average precision (AP) \cite{M-Zhu-TR-Waterloo-2004, M-Guillaumin-et-al-CVPR-2010}. In this paper, AP is the ranking performance computed under each label. Usually, the mean value over all labels, i.e., mAP, is reported.

In both of the datasets, twenty percent of the test images are used for validation, with the parameters performing best on the validation set used for testing. In all the following experiments, five random choices of the labeled instances are used, and the average performance with standard deviations is reported.

\begin{figure*}
\centering
\subfigure{\includegraphics[width=0.65\columnwidth]{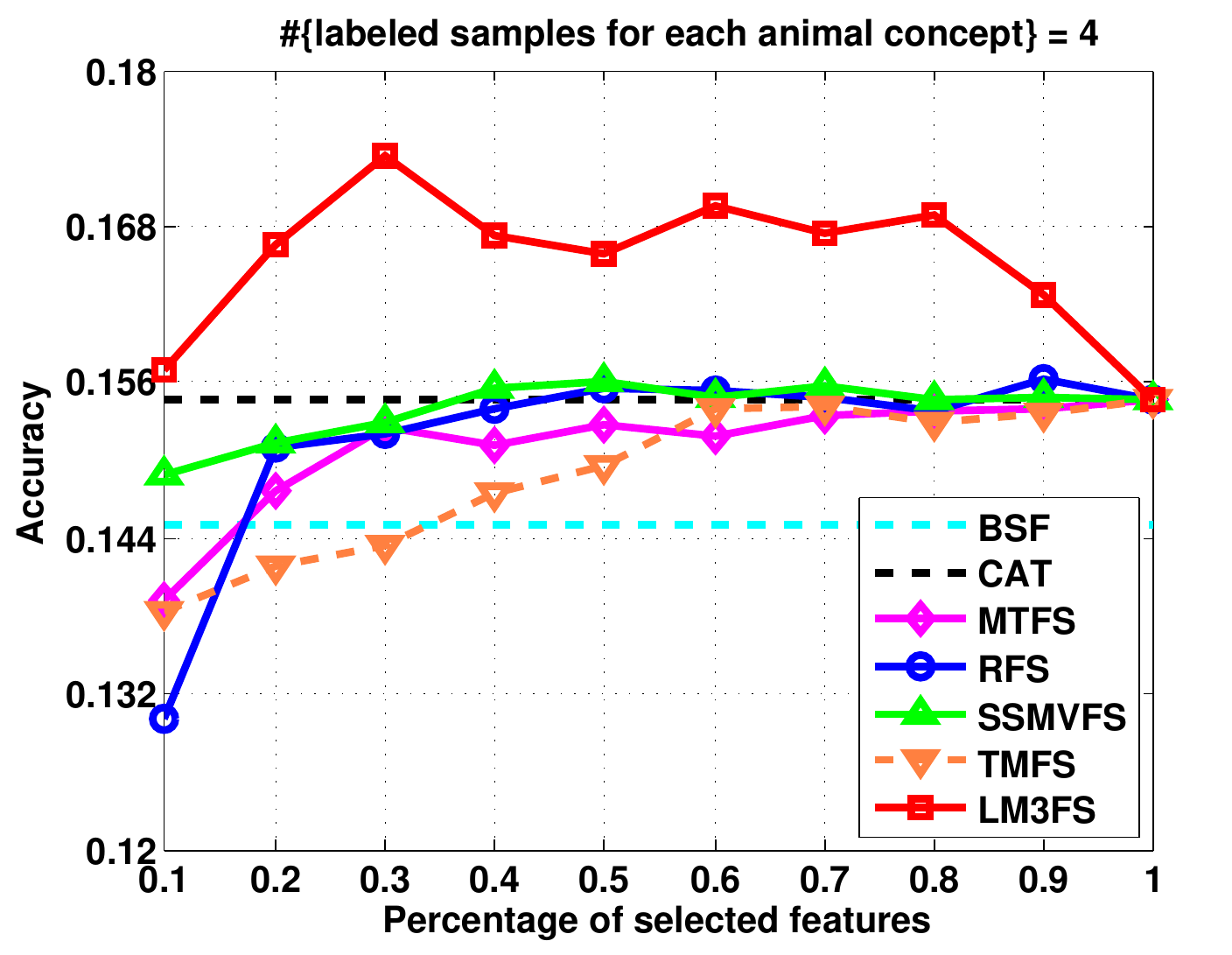}
}
\hfil
\subfigure{\includegraphics[width=0.65\columnwidth]{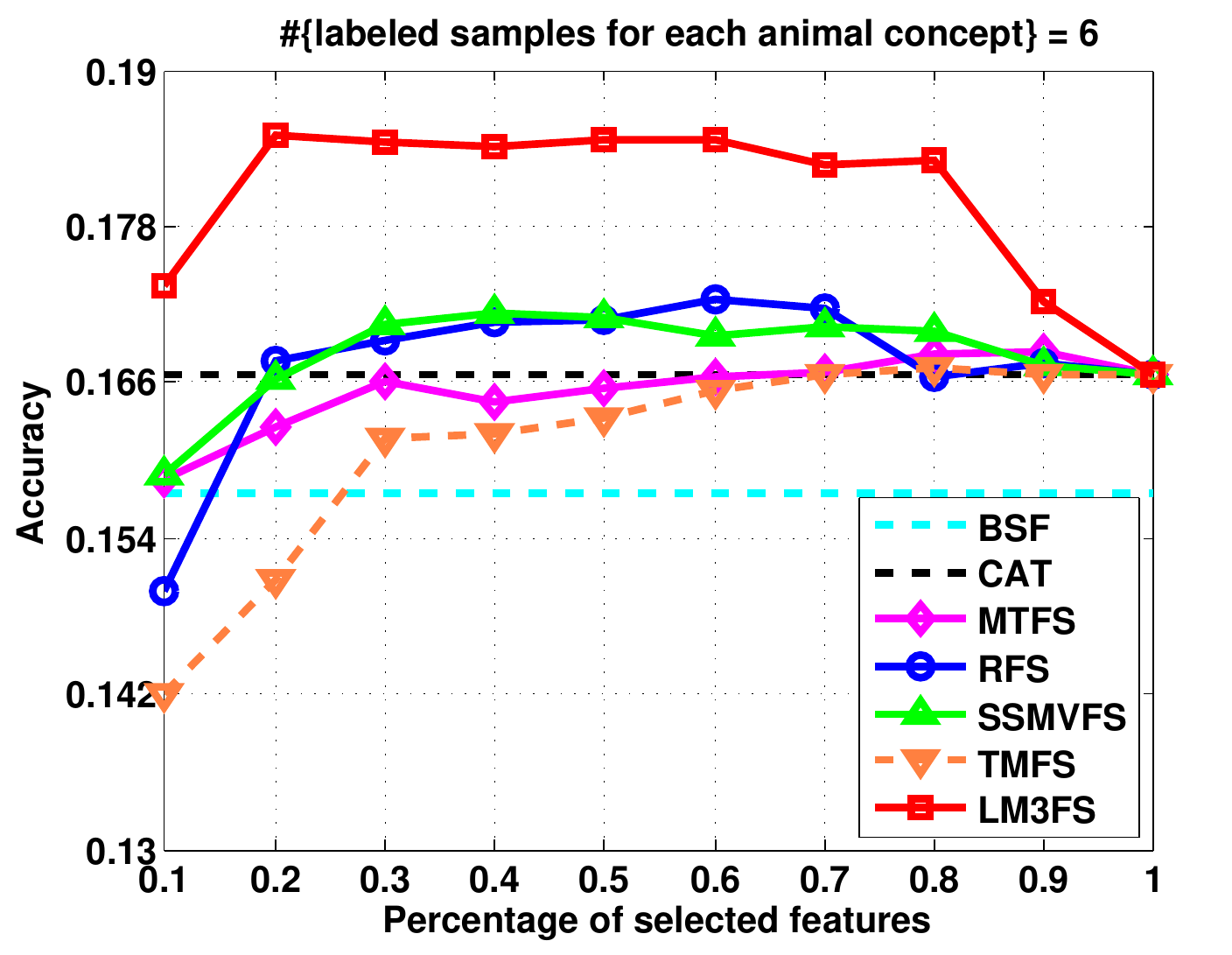}
}
\hfil
\subfigure{\includegraphics[width=0.65\columnwidth]{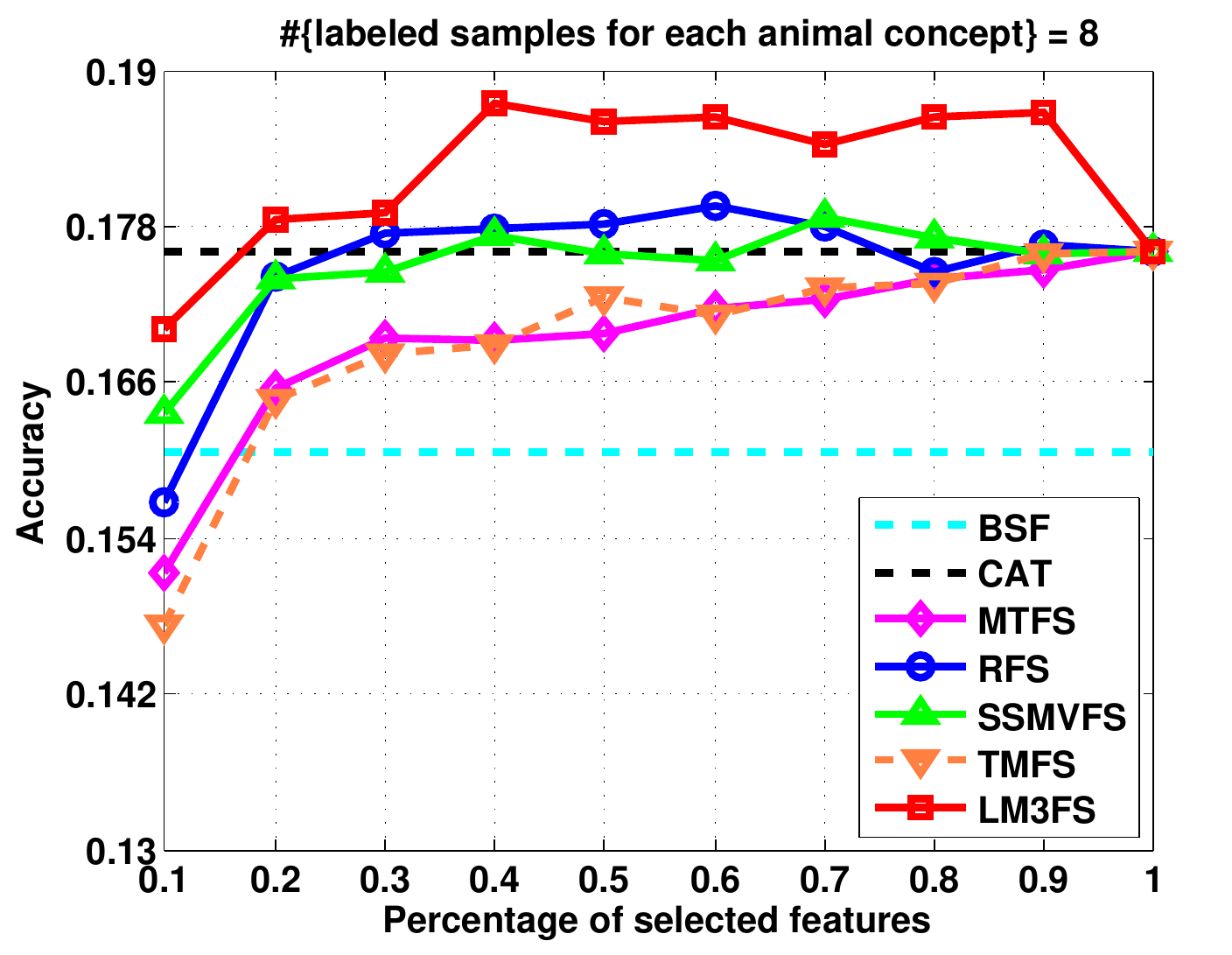}
}
\caption{Accuracies of the compared feature selection methods vs. percent of the original features being selected on the NUS-WIDE animal subset.}
\label{fig:Acc_vs_Dim_FS_NUS}
\end{figure*}

\begin{figure*}
\centering
\subfigure{\includegraphics[width=0.65\columnwidth]{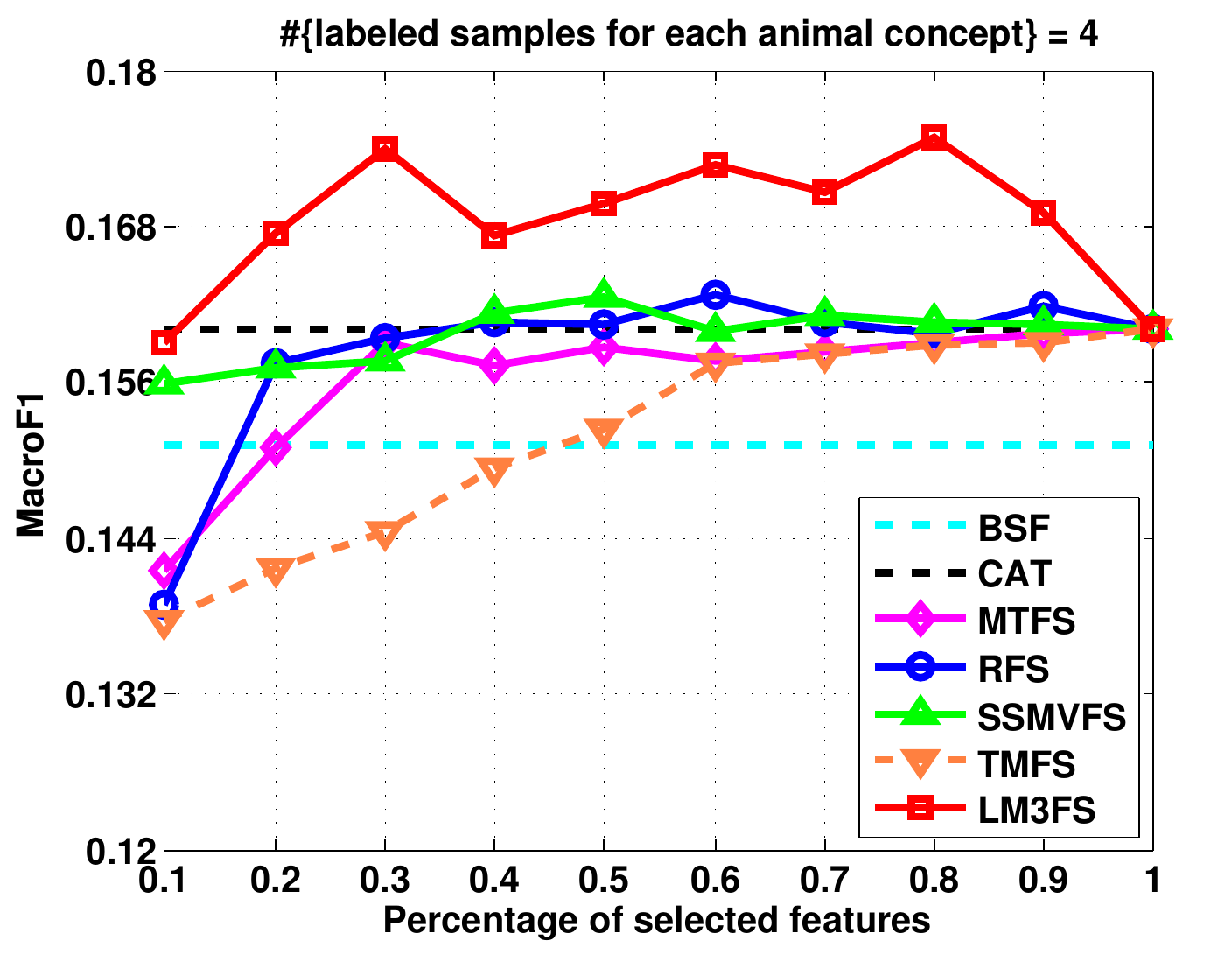}
}
\hfil
\subfigure{\includegraphics[width=0.65\columnwidth]{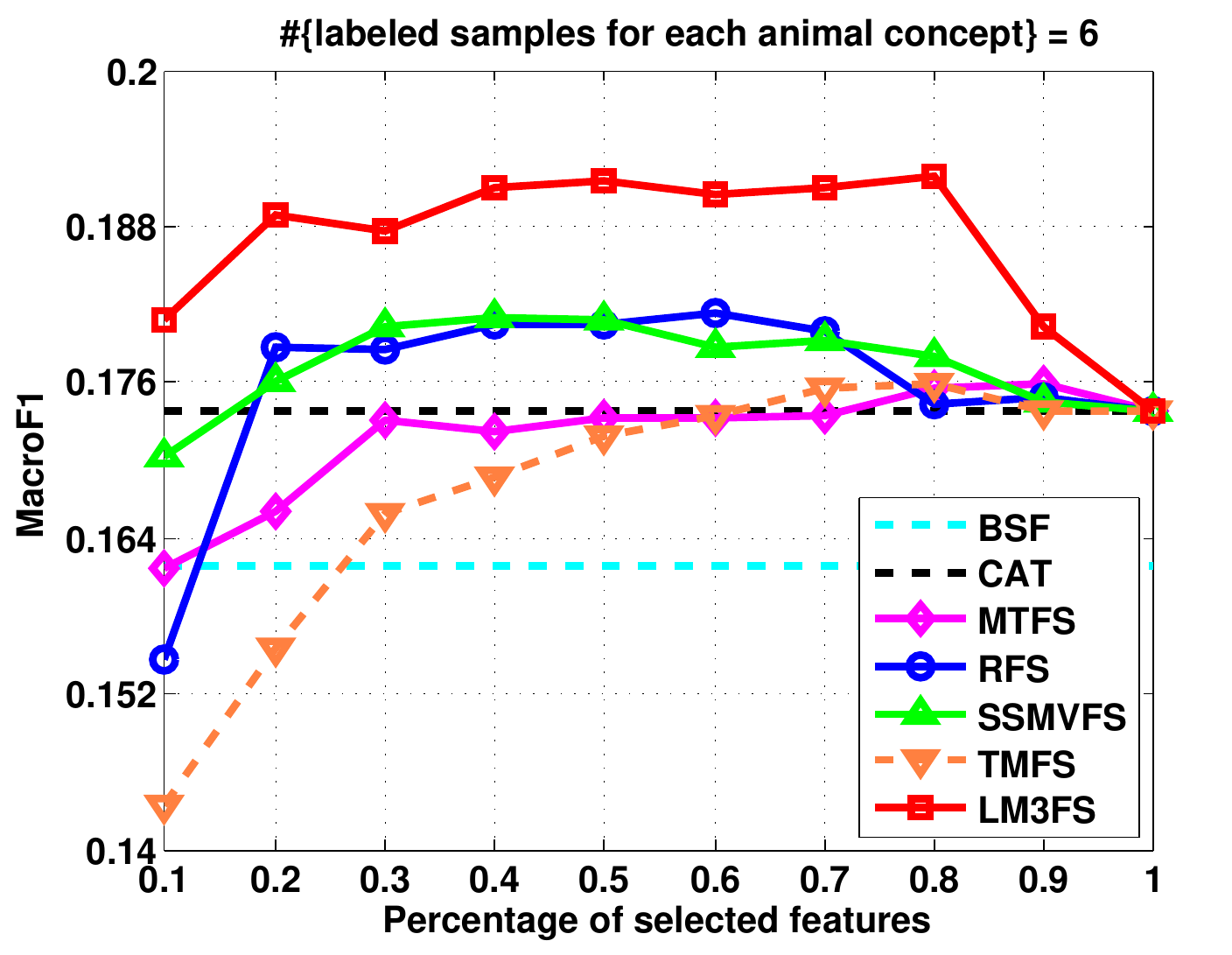}
}
\hfil
\subfigure{\includegraphics[width=0.65\columnwidth]{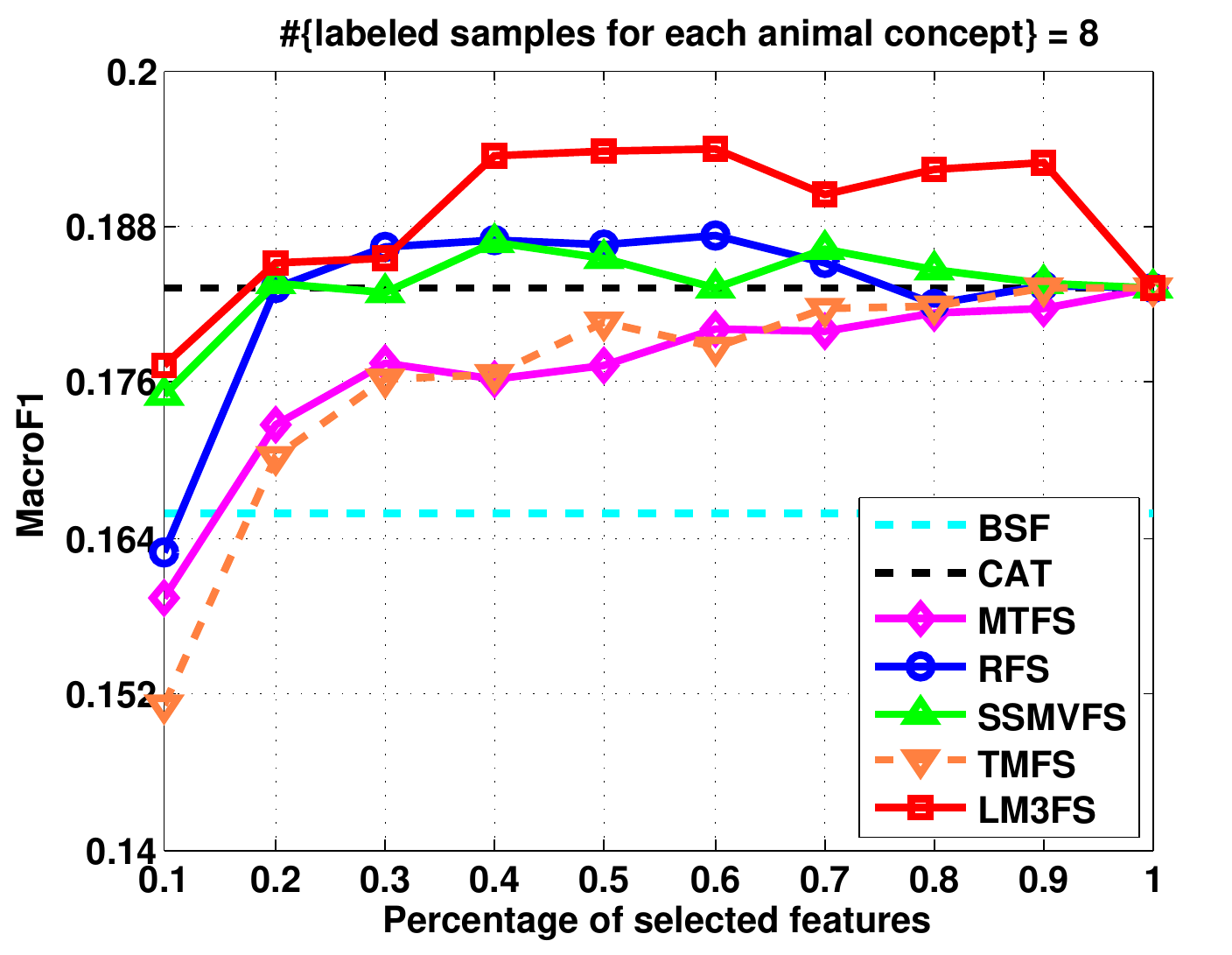}
}
\caption{MacroF1 scores of the compared feature selection methods vs. percent of the original features being selected on the NUS-WIDE animal subset.}
\label{fig:MacroF1_vs_Dim_FS_NUS}
\end{figure*}

\begin{figure*}
\centering
\subfigure{\includegraphics[width=0.65\columnwidth]{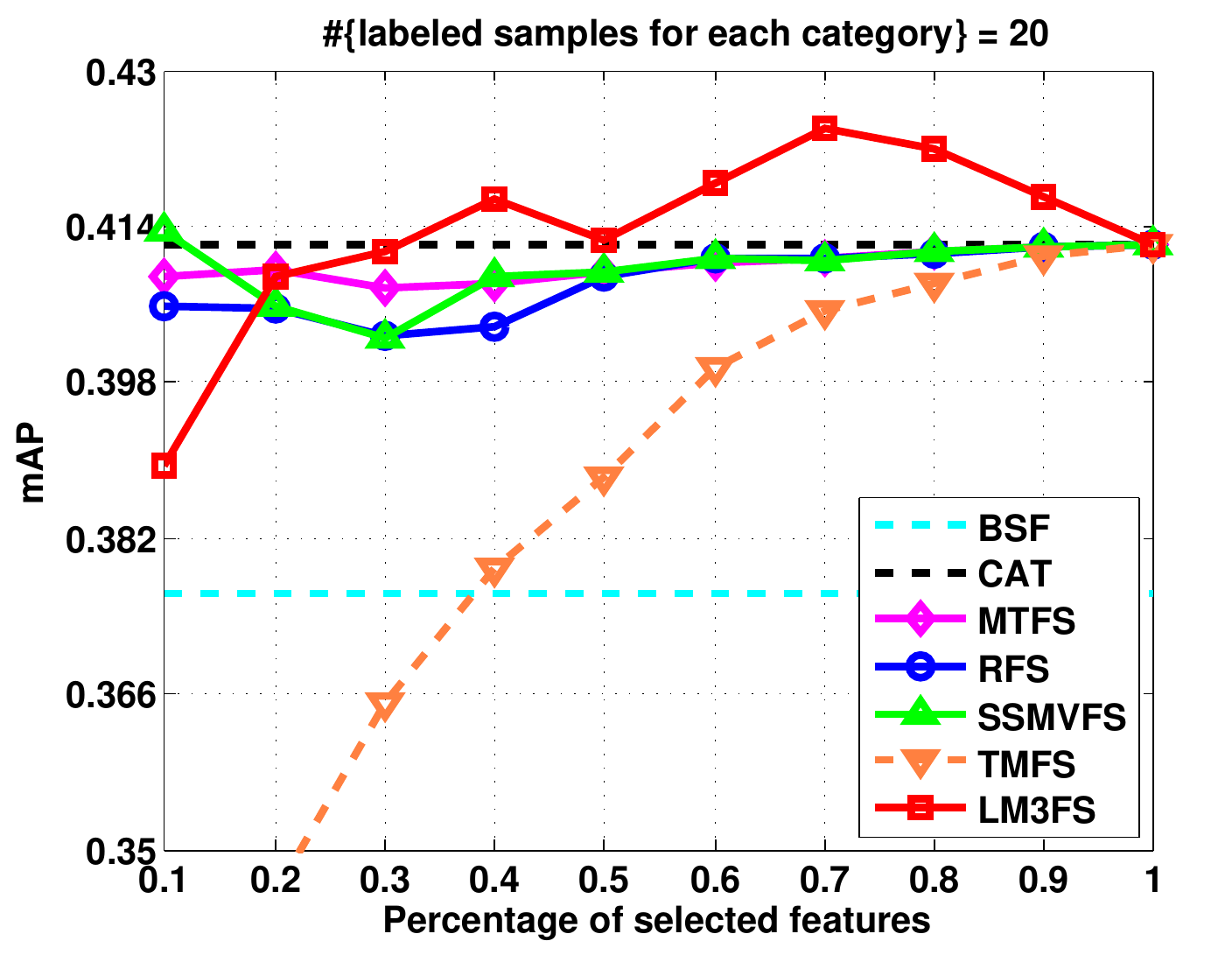}
}
\hfil
\subfigure{\includegraphics[width=0.65\columnwidth]{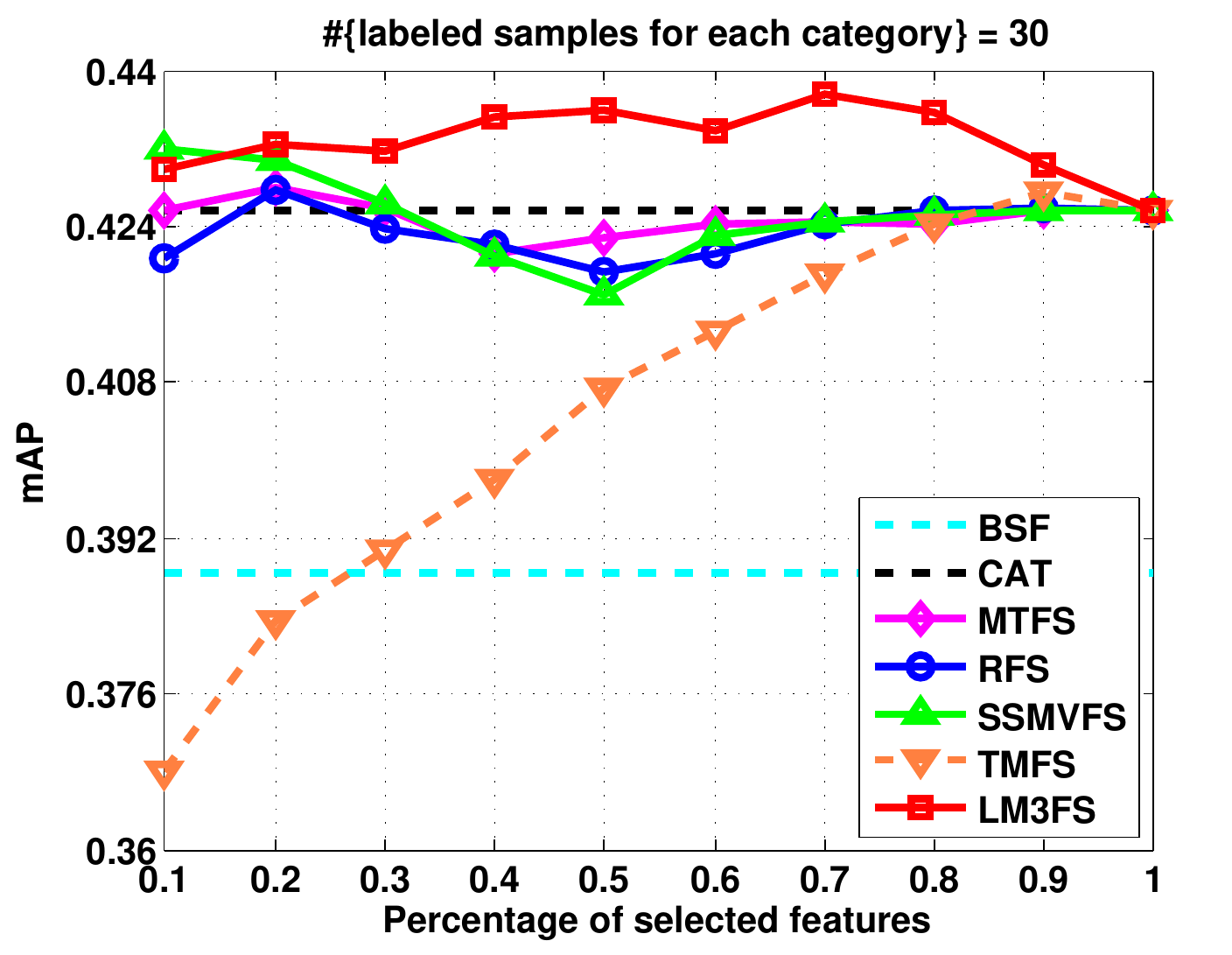}
}
\hfil
\subfigure{\includegraphics[width=0.65\columnwidth]{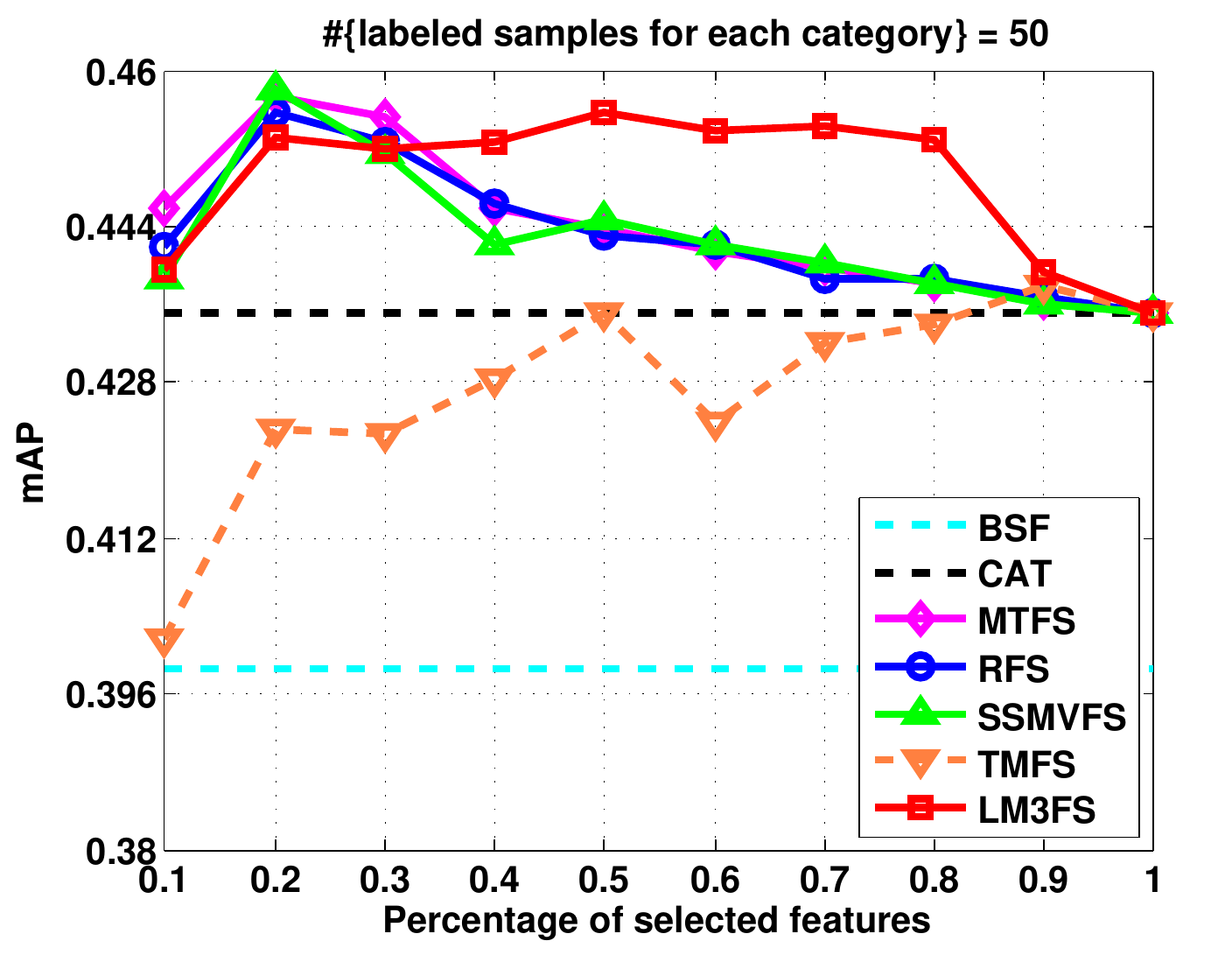}
}
\caption{mAPs of the compared feature selection methods vs. percent of the original features being selected on the MIR Flickr dataset.}
\label{fig:mAP_vs_Dim_FS_MIR}
\end{figure*}

\subsection{Feature selection evaluation}
\label{subsec:FS_Evaluation}

In this set of experiments, we use the obtained solutions of $\{ U^{(v)} \}$ for feature selection. For the $v$'th modality, the features are sorted in descending order according to the values $\| u^{(v),i} \|_2, i = 1,\ldots,d_v$, and then the $r_v$ top-ranked features are selected. The features are assumed to have been normalized and the selected features of all modalities are concatenated as the input of a subsequent classifier. Specially, we compare the following methods:
\begin{itemize}
  \item \textbf{BSF:} using the single-modal feature that achieves the best performance in $1$NN/RLS-based classification.
  \item \textbf{CAT:} concatenating the normalized features of all the modalities into a long vector, and then performing $1$NN/RLS-based classification.
  \item \textbf{MTFS \cite{G-Obozinski-et-al-ICMLw-2006}:} a supervised multi-task feature selection algorithm, where the formulation is given by (\ref{eq:MTFS_Formulation}) and the trade-off parameter $\gamma$ is set in the range $\{10^i | i=-5,-4,\ldots,5\}$.
  \item \textbf{RFS \cite{FP-Nie-et-al-NIPS-2010}:} an efficient and robust supervised multi-task feature selection algorithm that utilizes the $l_{2,1}$-norm for both the least squares loss and the regularization term. The formulation is given by (\ref{eq:RFS_Formulation}) and the trade-off parameter $\gamma$ is chosen from the set $\{ 10^i | i=-5,-4,\ldots,5 \}$.
  \item \textbf{SSMVFS \cite{H-Wang-et-al-ICML-2013}:} a competitive multi-modal multi-task feature learning algorithm that utilizes structured sparsity to explore the interrelations between multi-modal features. The two trade-off parameters $\gamma_1$ and $\gamma_2$ are tuned on the grid $\{ 10^i | i=-5,-4,\ldots,5 \}$ following \cite{H-Wang-et-al-ICML-2013}.
  \item \textbf{TMFS \cite{BK-Cao-et-al-ICDM-2014}:} a recently proposed multi-modal feature selection algorithm that utilizes the tensor-product to leverage the underlying multi-modal correlations. It eliminates features with the smallest ranking value one at a time and thus the task relationships are ignored. This method was originally designed for binary classification, and we extend it for the multi-class problem using the ``one-vs-all'' strategy, where multiple binary problems are created. The feature weights obtained from these binary problems are averaged to compute the final ranking value for each feature.
  \item \textbf{LM3FS:} the proposed multi-modal feature selection method. The candidate set for the parameters $\gamma_B$ is $\{ 10^i | i=-9,-8,\ldots,1 \}$, and $\gamma_A$, $\gamma_C$ are optimized over the set $\{ 10^i | i=-5,-4,\ldots,5 \}$.
\end{itemize}

In MTFS, RFS, and SSMVFS, the feature weight (or selection) matrices $\{ U^{(v)} \}$ are learned by concatenating them as a single matrix $U$. The features are selected according to $U^{(v)}$ for the $v$'th modality, and the number of selected features $r_v$ varies in $\{ 0.1,0.2,\ldots,1.0 \}$ of the original feature dimensions. The accuracies and macroF1 scores on the NUS animal subset are shown in Fig. \ref{fig:Acc_vs_Dim_FS_NUS} and Fig. \ref{fig:MacroF1_vs_Dim_FS_NUS} respectively. From these results, we observe that: 1) the performance of all the compared methods improves with an increased number of labeled instances; 2) the simple concatenation strategy (CAT) is superior to the best single modality (BSF), and by feature selection, we obtain further improvements. In particular, by selecting only $20$ to $30$ percent of the original features, most of the selection methods obtain results comparable to or better than the use of all features; 3) the robust feature selection algorithm (RFS) is better than the traditional multi-task feature selection (MTFS) and seems to be comparable with the multi-modal feature selection approach (SSMVFS). This may be because SSMVFS is not particularly designed for classification, and thus weakly predictive features may be selected. The proposed LM3FS outperforms both of them significantly. This demonstrates the significance of selecting features that have strong prediction power for classification; 4) although TMFS is able to select predictive features, it is only comparable with MTFS. This may principally be that TMFS was originally developed for binary classification, and a more sophisticated algorithm than the simple strategy adopted here should be designed for multi-class classification. This also indicates that exploring the task (label) relationships is also critical for the feature selection of multiple concepts; 5) the performance under the accuracy and macroF1 score criteria are consistent.

We show the results on the MIR dataset in Fig. \ref{fig:mAP_vs_Dim_FS_MIR}. The main differences from the results on the NUS dataset are that: 1) all MTFS, RFS, and SSMVFS are comparable to each other, and significantly superior to TMFS. This is because MIR is a multi-label dataset, and most images have more than one class label. This increases the significance of the label relationships in feature selection, thus the multi-task feature selection algorithms tend to be very competitive. For example, when the number of labeled samples for each category is $50$, the best performance (peak of the curve) of all other multi-task based approaches is comparable to our method; 2) the dimensions (percentage shown here) that achieve the best performance are smaller than those on the NUS dataset. The main reason is that the bag of SIFT visual words (BOVW) and tags utilized are both very sparse features, which are often more suitable for feature selection or transformation than the compact features. Although BOVW is also used in the NUS dataset, the dimension is smaller, thus the features tend to be more compact.

\subsection{Feature transformation evaluation}
\label{subsec:FT_Evaluation}

This set of experiments is conducted by applying the obtained solutions of $\{ U^{(v)} \}$ and $\{ \theta_v \}$ for feature transformation, i.e., the transformed representation $\sum_{v=1}^V \theta_v (U^{(v)})^T x_n^{(v)}$ is used for the subsequent classification. Most of the feature selection methods compared in Section \ref{subsec:FS_Evaluation} can also be used for feature transformation, with the exception of the TMFS algorithm, which eliminates features with the least predictive power one by one and does not learn extraction matrices. The other feature selection approaches that can be used for feature transformation are termed MTFT, RFT, SSMVFT, and LM3FT respectively in this section. The final feature dimension is equal to the number of class labels $P$, and the parameters are tuned in the same way as in the last set of experiments. To facilitate comparison, we also present the results of MTFS, RFS, SSMVFS, and LM3FS at the number of selected features (percentages) that achieve their best performance.

\begin{figure*}
\centering
\subfigure{\includegraphics[width=0.95\columnwidth]{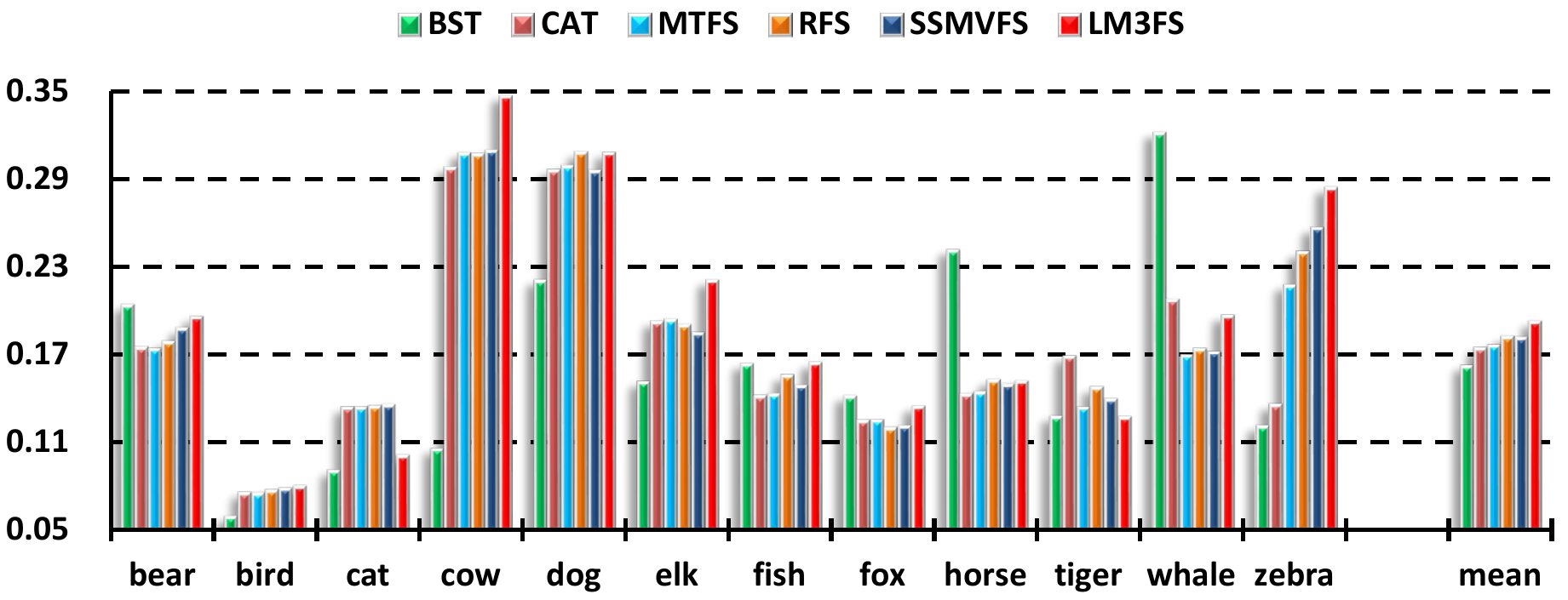}
}
\hfil
\subfigure{\includegraphics[width=0.95\columnwidth]{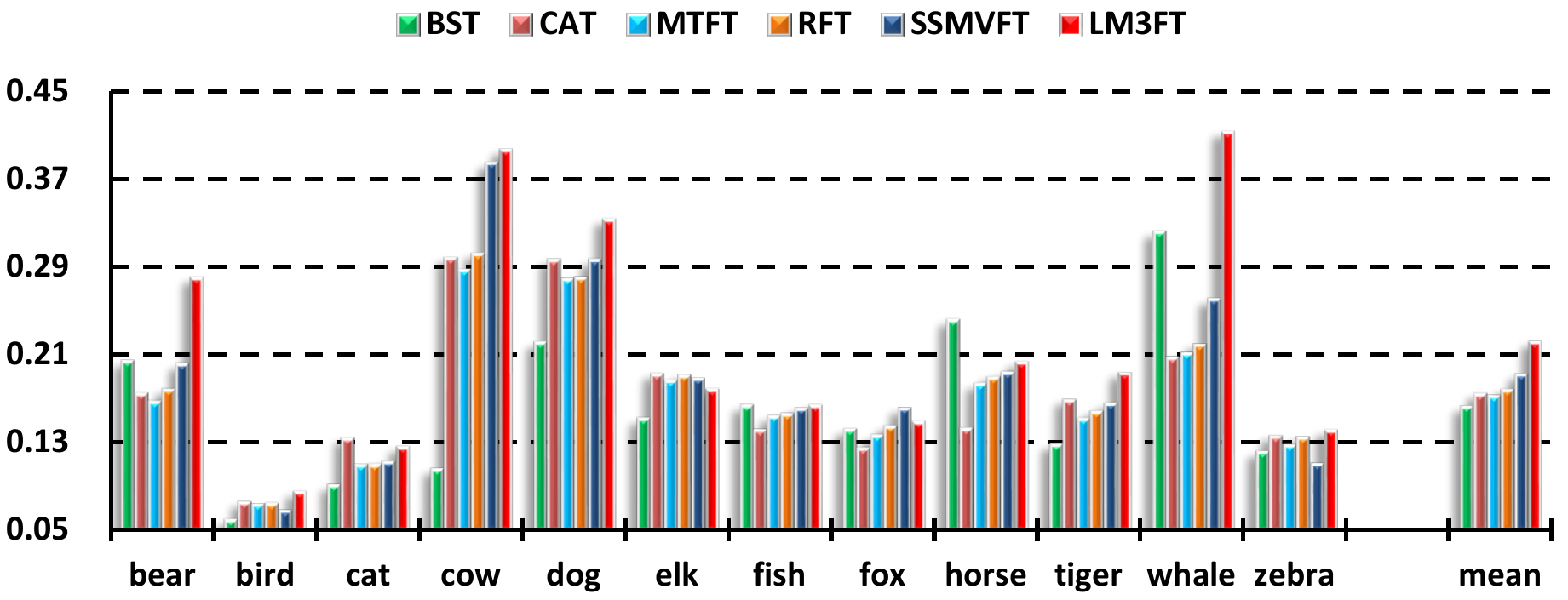}
}
\caption{F1 scores for the individual classes on the NUS-WIDE animal subset. 6 labeled samples for each animal concept are utilized. (Left: the feature selection methods at their best dimensions; Right: the feature transformation methods.)}
\label{fig:IndvF1_NUS}
\end{figure*}

We first show the F1 scores for the individual classes on the NUS-WIDE animal subset in Fig. \ref{fig:IndvF1_NUS}. From the results, we can see that most categories are hard to distinguish since they are similar to other categories (e.g.,  ``cat'' is similar to ``tiger''), or have large intra-class variability (such as ``bird''). Some categories are a bit easier since they have low similarity to other categories (such as ``bear'') or small intra-class variability (such as ``cow''). Overall, it is a challenge dataset due to its high inter-class similarity and large intra-class variability.

\begin{table*}[!t]
\setlength\tabcolsep{2pt}
\renewcommand{\arraystretch}{1.3}
\caption{Accuracies and MacroF1 scores of the compared feature transformation methods (as well as the feature selection methods at their best dimensions) on the NUS-WIDE animal subset.}
\label{tab:Acc_MacroF1_FS_FT_NUS}
\centering
\begin{tabular}{c||c|c|c||c|c|c}
\hline
\ & \multicolumn{3}{c||}{Accuracy} & \multicolumn{3}{c}{MacroF1} \\
\hline
Methods & 4 & 6 & 8 & 4 & 6 & 8 \\
\hline
BSF & 0.145$\pm$0.017 & 0.158$\pm$0.012 & 0.161$\pm$0.011 & 0.151$\pm$0.017 & 0.162$\pm$0.016 & 0.166$\pm$0.016 \\
\hline
CAT & 0.155$\pm$0.018 & 0.167$\pm$0.017 & 0.176$\pm$0.015 & 0.160$\pm$0.017 & 0.174$\pm$0.018 & 0.183$\pm$0.015 \\
\hline
MTFS (best dim) & 0.155$\pm$0.018 & 0.168$\pm$0.018 & 0.176$\pm$0.015 & 0.160$\pm$0.017 & 0.176$\pm$0.018 & 0.183$\pm$0.015 \\
RFS (best dim) & 0.156$\pm$0.018 & 0.172$\pm$0.017 & 0.180$\pm$0.018 & 0.163$\pm$0.018 & 0.181$\pm$0.018 & 0.187$\pm$0.018 \\
SSMVFS (best dim) & 0.156$\pm$0.024 & 0.171$\pm$0.019 & 0.179$\pm$0.016 & 0.163$\pm$0.021 & 0.181$\pm$0.020 & 0.187$\pm$0.015 \\
LM3FS (best dim) & 0.173$\pm$0.013 & 0.185$\pm$0.011 & 0.188$\pm$0.015 & 0.175$\pm$0.012 & 0.192$\pm$0.015 & 0.194$\pm$0.019 \\
\hline
MTFT & 0.160$\pm$0.019 & 0.170$\pm$0.020 & 0.174$\pm$0.015 & 0.161$\pm$0.017 & 0.171$\pm$0.019 & 0.177$\pm$0.015 \\
RFT & 0.163$\pm$0.021 & 0.177$\pm$0.019 & 0.179$\pm$0.013 & 0.165$\pm$0.019 & 0.178$\pm$0.016 & 0.180$\pm$0.013 \\
SSMVFT & 0.171$\pm$0.014 & 0.187$\pm$0.012 & 0.190$\pm$0.008 & 0.174$\pm$0.015 & 0.191$\pm$0.010 & 0.189$\pm$0.010 \\
LM3FT & \textbf{0.184$\pm$0.032} & \textbf{0.201$\pm$0.042} & \textbf{0.217$\pm$0.047} & \textbf{0.194$\pm$0.024} & \textbf{0.221$\pm$0.029} & \textbf{0.231$\pm$0.023} \\
\hline
\end{tabular}
\end{table*}

We also report the performance over all classes on the two datasets in Table \ref{tab:Acc_MacroF1_FS_FT_NUS} and Table \ref{tab:mAP_FS_FT_MIR} respectively. It can be seen from the results that: 1) the performance of the feature transformation strategy is better than the CAT baseline, as well as its corresponding feature selection strategy in most cases. This is mainly because the selection process may discard some useful information, which is usually preserved in transformation; 2) feature selection is more stable than transformation, since performance of the latter will be very unsatisfactory (e.g., RFT on the MIR dataset) if the learned transformation matrix is unreliable. This never occurs with feature selection, since it can never be worse than the CAT baseline if enough features are selected; 3) on the NUS dataset, SSMVFT is superior to both MTFT and RFT, which are comparable to the CAT baseline, and the proposed LM3FT significantly outperforms all the other transformation methods. Although SSMVFT is comparable to our method when $50$ labeled positive instances are utilized, the proposed LM3FT still achieves the best performance overall on the MIR dataset; 4) On the NUS dataset, the standard deviation of the proposed LM3FS is less than or comparable to the compared MTFS, RFS, and SSMVFS. However, the standard deviation of LM3FT is larger than the compared MTFT, RFT, and SSMVFT. This is mainly because the proposed method learns an additional large margin prediction parameter ``W'' compared with the other methods. Although it helps to approximate the true underlying model, a large variance (standard deviation) may be obtained according to the machine learning theory on bias and variance trade-off (\cite{C-Sammut-and-G-Webb-Springer-2011, G-James-et-al-Springer-2013}). We therefore add a regularization term $\|W\|_F^2$ to control the model complexity, so that the resulted standard deviation is tolerable for real world applications. In addition, the standard deviations of the proposed method are comparable to other approaches on the MIR dataset. This indicates a good bias and variance trade-off of the proposed model.

\section{Conclusion}
%The conclusion goes here.
This paper presents a large margin multi-modal multi-task feature extraction (LM3FE) framework. The framework simultaneously utilizes the information shared between tasks and the complementarity of different modalities to extract strongly predictive features for image classification. The framework can either be used for feature selection (LM3FS) or as a feature transformation (LM3FT) method. We investigated these two strategies and experimentally compared them with other multi-modal feature extraction approaches. We mainly conclude that: 1) both the label relationships and modality correlations are critical for multi-modal feature selection in multi-class or multi-label image classification, and by selecting features with strongly predictive power we can usually obtain significant improvements; 2) sparse features seem to be more appropriate for feature selection or transformation than compact features; and 3) it seems that feature transformation is better than feature selection when the same feature weight matrix is used, but the performance of feature selection tends to be more stable.

\begin{table}[!t]
\setlength\tabcolsep{2pt}
\renewcommand{\arraystretch}{1.3}
\caption{mAPs of the compared feature transformation methods (as well as the feature selection methods at their best dimensions) on the MIR Flickr dataset.}
\label{tab:mAP_FS_FT_MIR}
\centering
\begin{tabular}{c||c|c|c}
\hline
Methods & 20 & 30 & 50 \\
\hline
BSF & 0.376$\pm$0.003 & 0.388$\pm$0.002 & 0.400$\pm$0.002 \\
\hline
CAT & 0.412$\pm$0.003 & 0.426$\pm$0.002 & 0.435$\pm$0.003 \\
\hline
MTFS (best dim) & 0.412$\pm$0.003 & 0.428$\pm$0.004 & 0.457$\pm$0.004 \\
RFS (best dim) & 0.412$\pm$0.003 & 0.428$\pm$0.004 & 0.456$\pm$0.004 \\
SSMVFS (best dim) & 0.413$\pm$0.005 & 0.432$\pm$0.004 & 0.458$\pm$0.004 \\
LM3FS (best dim) & 0.424$\pm$0.005 & 0.438$\pm$0.003 & 0.456$\pm$0.003 \\
\hline
MTFT & 0.424$\pm$0.004 & 0.438$\pm$0.003 & 0.451$\pm$0.002 \\
RFT & 0.386$\pm$0.004 & 0.394$\pm$0.007 & 0.402$\pm$0.007 \\
SSMVFT & 0.441$\pm$0.006 & 0.455$\pm$0.005 & \textbf{0.469$\pm$0.002} \\
LM3FT & \textbf{0.451$\pm$0.008} & \textbf{0.462$\pm$0.010} & \textbf{0.469$\pm$0.009} \\
\hline
\end{tabular}
\end{table}

% or


%\appendices

%\section{}

\appendices

\section{Proof of Theorem \ref{thm:Gradient_Lipschitz_Cst_g_wp}}
\begin{IEEEproof}
According to (\ref{eq:Smoothed_g}), (\ref{eq:PieceWise_g}) and the reformulation $\hbar_p(\{ x_n^{(v)} \}) = \hbar_p(z_n, w_p) = w_p^T z_n + b_p$, the gradient of $g^\sigma$ w.r.t. $w_p$ for the $n$'th sample is
\begin{equation}
\label{eq:Gradient_g_wp_xn_Derivation}
\begin{split}
& \frac{\partial g^\sigma(x_n^{(v)}, y_{pn}, w_p)}{\partial w_p} = \\
& \left\{
\begin{array}{cc}
  0, & \ \nu_{pn} = 0; \\
  -y_{pn} z_n, & \ \nu_{pn} = 1; \\
  \frac{(-y_{pn} z_n)(1 - y_{pn} \hbar_p(z_n, w_p))}{\sigma \|x_n\|_\infty}, & \ \nu_{pn} = \frac{1 - y_{pn} \hbar(z_n, w_p)}{\sigma \|x_n\|_\infty}.
\end{array}
\right.
\end{split}
\end{equation}
This indicates that
\begin{equation}
\label{eq:Gradient_g_wp_xn_Summary}
\frac{\partial g^\sigma(x_n^{(v)}, y_{pn}, w_p)}{\partial w_p} = -y_{pn} z_n \nu_{pn}.
\end{equation}
Thus the sum of the gradient over all the $N$ samples is
\begin{equation}
\label{eq:Gradient_g_wp_Derivation}
\frac{\partial g^\sigma(w_p)}{\partial w_p} = \frac{\partial \sum_{n=1}^N g^\sigma(z_n, y_{pn}, w_p)}{\partial w_p} = - Z Y_p \nu_p.
\end{equation}
Given function $g(x)$, for any $x^{(1)}$ and $x^{(2)}$, the Lipschitz constant $L$ satisfies
\begin{equation}
\label{eq:Lipschitz_g_Definition}
\|\nabla g(x^{(1)}) - \nabla g(x^{(2)})\|_2 \leq L \|x^{(1)} - x^{(2)}\|_2.
\end{equation}
Hence the Lipschitz constant of $g^\sigma$ w.r.t. $w_p$ can be calculated from
\begin{equation}
\mathrm{max} \frac{\left\|\frac{\partial g^\sigma}{\partial w_p^{(1)}} - \frac{\partial g^\sigma}{\partial w_p^{(2)}}\right\|_2}{\|w_p^{(1)} - w_p^{(2)}\|_2} \leq L_g^\sigma(z_n, y_{pn}, w_p).
\end{equation}
According to (\ref{eq:Gradient_g_wp_xn_Derivation}), we have
\begin{equation}
\begin{split}
& \frac{\partial g^\sigma}{\partial w_p^{(1)}} - \frac{\partial g^\sigma}{\partial w_p^{(2)}} \\
& = \left\{
\begin{array}{cc}
  0, & y_{pn} \hbar_p(x_n) > 1 \ \mathrm{or} \ < 1 - \sigma \|x_n\|_\infty; \\
  \frac{z_n z_n^T (w_p^{(1)} - w_p^{(2)})}{\sigma \|x_n\|_\infty}, & \mathrm{else}.
\end{array}
\right.
\end{split}
\end{equation}
Therefore,
\begin{equation}
\begin{split}
\mathrm{max} & \frac{\|z_n z_n^T (w_p^{(1)} - w_p^{(2)})\|_2}{\sigma \|x_n\|_\infty \|w_p^{(1)} - w_p^{(2)}\|_2} \leq \frac{\|z_n z_n^T\|_2}{\sigma \|x_n\|_\infty} = L^g(z_n, y_{pn}, w_p).
\end{split}
\end{equation}
To this end, the Lipschitz constant of $g^\sigma(w_p)$ is calculated as
\begin{equation}
\begin{split}
\sum_n L^g(z_n, y_{pn}, w_p) & \leq N \mathrm{max}_n L^g(z_n, y_{pn}, w_p) \\
& = \frac{N}{\sigma} \mathrm{max}_n \frac{\|z_n z_n^T\|_2}{\|x_n\|_\infty} = L_g^\sigma(w_p).
\end{split}
\end{equation}
This completes the proof.
\end{IEEEproof}

\section{Proof of Theorem \ref{thm:Gradient_Lipschitz_Cst_Uv}}
\begin{IEEEproof}
According to (\ref{eq:Smoothed_g}), (\ref{eq:PieceWise_g}) and the reformulation $\hbar_p(\{ x_n^{(v)} \}) = \hbar_p(x_n^{(v)}, U^{(v)}) = \theta_v w_p^T (U^{(v)})^T x_n^{(v)} + c_{pn}^{(v)}$, the gradient of $g^\sigma$ w.r.t. $U^{(v)}$ for the $n$'th sample is
\begin{equation}
\label{eq:Gradient_g_Uv_xn_Derivation}
\begin{split}
& \frac{\partial g^\sigma(x_n^{(v)}, y_{pn}, U^{(v)})}{\partial U^{(v)}} = \\
& \left\{
\begin{array}{cc}
  0, & \ \nu_{pn} = 0; \\
  -y_{pn} \theta_v x_n^{(v)} w_p^T, & \ \nu_{pn} = 1; \\
  \frac{(-y_{pn} \theta_v x_n^{(v)} w_p^T)(1 - y_{pn} \hbar_p(x_n^{(v)}, U^{(v)}))}{\sigma \|x_n\|_\infty}, & \nu_{pn} = \frac{1 - y_{pn} \hbar(x_n^{(v)}, U^{(v)})}{\sigma \|x_n\|_\infty}.
\end{array}
\right.
\end{split}
\end{equation}
This indicates that
\begin{equation}
\label{eq:Gradient_g_Uv_xn_Summary}
\frac{\partial g^\sigma(x_n^{(v)}, y_{pn}, U^{(v)})}{\partial U^{(v)}} = (-y_{pn} \theta_v x_n^{(v)} w_p^T) \nu_{pn}.
\end{equation}
Thus the sum of the gradient over all the $N$ samples and $P$ labels is
\begin{equation}
\label{eq:Gradient_g_Uv_Derivation}
\begin{split}
\frac{\partial g^\sigma(U^{(v)})}{\partial U^{(v)}} & = \frac{\partial \sum_{p=1}^P \sum_{n=1}^N g^\sigma(x_n^{(v)}, y_{pn}, U^{(v)})}{\partial U^{(v)}} \\
& = \sum_{p=1}^P - \theta_v X^{(v)} Y_p \nu_p w_p^T.
\end{split}
\end{equation}
According to (\ref{eq:Lipschitz_g_Definition}), the Lipschitz constant of $g^\sigma$ w.r.t. $U^{(v)}$ can be calculated from
\begin{equation}
\mathrm{max} \frac{\left\|\frac{\partial g^\sigma}{\partial U^{(v),(1)}} - \frac{\partial g^\sigma}{\partial U^{(v),(2)}}\right\|_2}{\|U^{(v),(1)} - U^{(v),(2)}\|_2} \leq L_g^\sigma(x_n^{(v)}, y_{pn}, U^{(v)}).
\end{equation}
According to (\ref{eq:Gradient_g_Uv_xn_Derivation}), we have
\begin{equation}
\begin{split}
& \frac{\partial g_p^\sigma}{\partial U^{(v),(1)}} - \frac{\partial g_p^\sigma}{\partial U^{(v),(2)}} = \\
& \left\{
\begin{array}{cc}
  0, & \begin{array}{c}
        y_{pn} \hbar_p(x_n) > 1 \ \mathrm{or}\\
        < 1 - \sigma \|x_n\|_\infty;
      \end{array} \\
  \frac{(\theta_v^2 x_n^{(v)} w_p^T)[(x_n^{(v)})^T (U^{(v),(1)} - U^{(v),(2)}) w_p]}{\sigma \|x_n\|_\infty}, & \mathrm{else}.
\end{array}
\right.
\end{split}
\end{equation}
Therefore,
\begin{equation}
\begin{split}
\mathrm{max} & \frac{\|(\theta_v^2 x_n^{(v)} w_p^T)[(x_n^{(v)})^T (U^{(v),(1)} - U^{(v),(2)}) w_p]\|_2}{\sigma \|x_n\|_\infty \|U^{(v),(1)} - U^{(v),(2)}\|_2} \\
& \leq \frac{\theta_v^2 \|x_n^{(v)} w_p^T\|_2 \|(x_n^{(v)})^T\|_2 \|w_p\|_2}{\sigma \|x_n\|_\infty} = L_g^\sigma(x_n^{(v)}, y_{pn}, U^{(v)}).
\end{split}
\end{equation}
To this end, the Lipschitz constant of $g_p^\sigma(U^{(v)})$ is calculated as
\begin{equation}
\begin{split}
& \sum_p \sum_n L^g(x_n^{(v)}, y_{pn}, U^{(v)}) \leq P N \mathop{\mathrm{max}}_p \mathop{\mathrm{max}}_n L_g^\sigma(x_n^{(v)}, y_{pn}, U^{(v)}) \\
& = \frac{P N \theta_v^2}{\sigma} \mathop{\mathrm{max}}_p \mathop{\mathrm{max}}_n \frac{\|x_n^{(v)} w_p^T\|_2 \|(x_n^{(v)})^T\|_2 \|w_p\|_2}{\|x_n\|_\infty} = L_g^\sigma(U^{(v)}).
\end{split}
\end{equation}
This completes the proof.
\end{IEEEproof}

\section{Proof of Theorem \ref{alg:Optimization_LM3FE}}
\begin{IEEEproof}
It can easily be verified that (\ref{eq:Gradient_F_Uv}) and (\ref{eq:Lipschitz_Cst_F_Uv}) are the gradient and Lipschitz constant of the following problem:
\begin{equation}
\label{eq:Equivalent_Formulation_wrt_Uv}
\mathop{\mathrm{argmin}}_{U^{(v)}} \ \Phi(U^{(v)}) + \gamma_B \mathrm{tr}\left( (U^{(v)})^T D^{(v)} U^{(v)} \right).
\end{equation}
Thus $U_{t+1}^{(v)}$ in (\ref{eq:Form_Uv}) is the solution of (\ref{eq:Equivalent_Formulation_wrt_Uv}) after the $t$'th iteration round if the Nesterov optimal gradient method is applied. This indicates that
\begin{equation}
\begin{split}
& \Phi(U_{t+1}^{(v)}) + \gamma_B \mathrm{tr}\left( (U_{t+1}^{(v)})^T D_t^{(v)} U_{t+1}^{(v)} \right) \\
\leq & \Phi(U_t^{(v)}) + \gamma_B \mathrm{tr}\left( (U_t^{(v)})^T D_t^{(v)} U_t^{(v)} \right),
\end{split}
\end{equation}
That is to say,
\begin{equation}
\Phi(U_{t+1}^{(v)}) + \gamma_B \sum_i \frac{\|u_{t+1}^{(v),i}\|_2^2}{2\|u_t^{(v),i}\|_2} \leq \Phi(U_{t}^{(v)}) + \gamma_B \sum_i \frac{\|u_t^{(v),i}\|_2^2}{2\|u_t^{(v),i}\|_2}.
\end{equation}
Using a simple trick (simultaneously adding and subtracting a term), we have
\begin{equation}
\begin{split}
& \Phi(U_{t+1}^{(v)}) + \gamma_B \|U_{t+1}^{(v)}\|_{2,1} - \gamma_B \left( \|U_{t+1}^{(v)}\|_{2,1} - \sum_i \frac{\|u_{t+1}^{(v),i}\|_2^2}{2\|u_t^{(v),i}\|_2} \right) \\
& \leq \Phi(U_{t}^{(v)}) + \gamma_B \|U_t^{(v)}\|_{2,1} - \gamma_B \left( \|U_t^{(v)}\|_{2,1} - \sum_i \frac{\|u_t^{(v),i}\|_2^2}{2\|u_t^{(v),i}\|_2} \right).
\end{split}
\end{equation}
According to Lemma 1 presented in \cite{FP-Nie-et-al-NIPS-2010}, we have
\begin{equation}
\|U_{t+1}^{(v)}\|_{2,1} - \sum_i \frac{\|u_{t+1}^{(v),i}\|_2^2}{2\|u_t^{(v),i}\|_2} \leq \|U_t^{(v)}\|_{2,1} - \sum_i \frac{\|u_t^{(v),i}\|_2^2}{2\|u_t^{(v),i}\|_2}.
\end{equation}
Thus we obtain
\begin{equation}
\Phi(U_{t+1}^{(v)}) + \gamma_B \|U_{t+1}^{(v)}\|_{2,1} \leq \Phi(U_{t}^{(v)}) + \gamma_B \|U_t^{(v)}\|_{2,1}.
\end{equation}
This completes the proof.
\end{IEEEproof}

%\section*{Acknowledgment}

\ifCLASSOPTIONcaptionsoff
  \newpage
\fi



\bibliographystyle{IEEEtran}
\bibliography{./TIP-13616-2015}
\end{document}